\documentclass[10pt,journal, compsoc]{IEEEtran}

\ifCLASSOPTIONcompsoc
  \usepackage[nocompress]{cite}
\else
  \usepackage{cite}
\fi
\ifCLASSINFOpdf
  \usepackage[pdftex]{graphicx}
  \graphicspath{{../pdf/}{../jpeg/}}
  \DeclareGraphicsExtensions{.pdf,.jpeg,.png}
\else
\fi
\usepackage{amsmath}
\newcommand\norm[1]{\left\lVert#1\right\rVert}
\newcommand{\etal}{\mbox{\emph{et al.\ }}}

\usepackage{enumitem}
\usepackage{amssymb}

\usepackage{graphicx}
\usepackage{subcaption}
\usepackage{url}

\usepackage{hyperref}

\begin{document}
%
\title{Face Recognition via Centralized Coordinate Learning}
%
%
%
%


\author{Xianbiao Qi,
 ~Lei Zhang
\thanks{Xianbiao Qi and Lei Zhang are with the Department
of Computing, the Hong Kong Polytechnic University, Hong Kong,
China. Contact e-mails: (qixianbiao@gmail.com, cslzhang@comp.polyu.edu.hk).}
}

\IEEEtitleabstractindextext{%
\begin{abstract}

Owe to the rapid development of deep neural network (DNN) techniques and the emergence of large scale face databases, face recognition has achieved a great success in recent years. During the training process of DNN, the face features and classification vectors to be learned will interact with each other, while the distribution of face features will largely affect the convergence status of network and the face similarity computing in test stage. In this work, we formulate jointly the learning of face features and classification vectors, and propose a simple yet effective centralized coordinate learning (CCL) method, which enforces the features to be dispersedly spanned in the coordinate space while ensuring the classification vectors to lie on a hypersphere. An adaptive angular margin is further proposed to enhance the discrimination capability of face features. Extensive experiments are conducted on six face benchmarks, including those have large age gap and hard negative samples. Trained only on the small-scale CASIA Webface dataset with 460K face images from about 10K subjects, our CCL model demonstrates high effectiveness and generality, showing consistently competitive performance across all the six benchmark databases.

\end{abstract}



}

\maketitle

\IEEEdisplaynontitleabstractindextext

%
\IEEEpeerreviewmaketitle

\section{Introduction}
Face recognition~\cite{turk1991face, zhao2003face, ahonen2006face, simonyan2013fisher,lu2015surpassing, taigman2014deepface,  sun2014deep1, sun2014deep2, szegedy2015going, wen2016discriminative,sun2016sparsifying, liu2017sphereface} has a broad range of applications in our daily life, including access control, video surveillance, public safety, online payment, image search and family photo album management. As a classical yet active topic, face recognition has been extensively studied. One essential problem in face recognition is how to obtain discriminative face features. In the past, handcrafted local image descriptors, such as SIFT~\cite{lowe2004distinctive}, HOG~\cite{dalal2005histograms}, LBP~\cite{ojala2002multiresolution} and its variants~\cite{zhang2007histogram, liu2017local}, have been widely used to extract local face features. Meanwhile, Principal Component Analysis (PCA)~\cite{turk1991face}, Linear Discriminative Analysis (LDA)~\cite{belhumeur1997eigenfaces}, and Sparse Representation (SR)~\cite{wright2009robust, wagner2012toward, yang2011robust, deng2012extended, zhang2011sparse} are also popular methods to construct a global face representation.

Face recognition recently has made a breakthrough due to the rapid advancement of deep neural network (DNN) techniques, especially convolutional neural networks (CNNs)~\cite{lecun1998gradient, krizhevsky2012imagenet, simonyan2014very, szegedy2015going, lecun2015deep, he2016deep, szegedy2017inception}, and the availability of many large scale face databases~\cite{huang2008labeled, wolf2011face, yi2014learning, guo2016ms, kemelmacher2016megaface, cao2017vggface2}. There are mainly two key issues in deep face recognition: designing effective network structure for face representation and constructing discriminative loss functions for feature learning. Some representative and successful deep CNNs for face recognition include DeepFace~\cite{taigman2014deepface}, DeepIDs~\cite{sun2014deep1, sun2014deep2, sun2015deeply, sun2015deepid3}, FaceNet~\cite{schroff2015facenet}, Deep FR~\cite{parkhi2015deep}, Center Loss~\cite{wen2016discriminative}, and SphereFace~\cite{liu2017sphereface}.

Given the network structure and optimization method, the design of loss functions largely determines the final recognition performance. Many discriminative loss functions~\cite{taigman2014deepface, sun2014deep1, sun2014deep2, sun2015deeply, sun2015deepid3, schroff2015facenet, wen2016discriminative, liu2017sphereface} have been proposed to provide effective supervision information for the DNN training. Based on the different focuses of loss design, these works can be roughly categorized into three groups:

\begin{itemize}
\item Methods focusing on the form of final face feature, such as $L_2$-constrainted face representation~\cite{ranjan2017l2}, $L_2$ Normface~\cite{wang2017normface}, and Coco loss~\cite{liu2017rethinking}.

\item Methods investigating the importance of classification vector, such as Large-margin softmax loss~\cite{liu2016large} and SphereFace~\cite{liu2017sphereface}.

\item Methods providing additional types of supervision information, such as Center loss~\cite{wen2016discriminative} and Triplet loss in FaceNet~\cite{schroff2015facenet}.  
\end{itemize}


During the training process of DNN, the face feature $\mathbf{x}$ and the classification vector $\mathbf{w}$ will interact with each other. The distribution of features obtained in one training stage will largely affect the output of classification vectors, which will in turn pull the samples of one specific person gathering together in one region of the coordinate space. Many previous works~\cite{ranjan2017l2, wang2017normface,liu2016large, liu2017rethinking, liu2017sphereface} focus on either the formulation of feature $\mathbf{x}$ or the formulation of classification vector $\mathbf{w}$, while they ignore the impact of the distribution of $\mathbf{x}$ on $\mathbf{w}$ as well as the final convergence of network. If in one training stage of DNN, most face features lie in a certain quadrant, the final features and their corresponding classification vectors will very likely converge to that quadrant. This will make the separation of different face subjects less effective.

In this paper, we consider simultaneously the formulations of feature $\mathbf{x}$ and classification vector $\mathbf{w}$. To make the faces of different subjects more separable, we argue that the learned face features should span dispersedly across the whole coordinate space centered on the origin, and the classification vectors should lie on a hypersphere manifold to make the cosine-similarity computation consistent in the training and test stages. To this end, we propose a simple yet effective method, namely centralized coordinate learning (CCL), where we centralize the face features $\mathbf{x}$ to the origin of the space.
As illustrated in Fig. 2(e), the centralization operator can make the features span more dispersedly across the whole coordinate space. Meanwhile, we introduce an adaptive angular margin (AAM) to further improve the face feature discrimination capability, as illustrated in Fig. 2(f). 


The proposed CCL method demonstrates high effectiveness and generality. Trained only on the CASIA Webface~\cite{yi2014learning}, which has 460K face images from 10K subjects, CCL consistently shows competitive results with state-of-the-arts on six face benchmarks, including Labeled Face in the Wild (LFW)~\cite{huang2008labeled}, Cross-Age LFW (CALFW)~\cite{zheng2017cross}, Cross-Age Celebrity Database (CACD)~\cite{chen2014cross}, Similar-Looking LFW (SLLFW)~\cite{deng2017fine}, YouTube Face (YTF)~\cite{wolf2011face} and MegaFace~\cite{kemelmacher2016megaface} datasets. Even without enforcing the proposed AAM operator, the learned CCL model still exhibits leading performance.




\section{Related Works}
In the study of DNN based face recognition, there are two key factors to the final performance: network structure and loss function. In the following, we briefly review the relevant works from these two aspects.

\subsection{Network Structure}
Motivated by the success of AlexNet~\cite{krizhevsky2012imagenet} in ImageNet competition~\cite{russakovsky2015imagenet}, researchers promptly adopted DNN in the field of face recognition. DeepFace~\cite{taigman2014deepface} and DeepIDs~\cite{sun2014deep1, sun2014deep2} are the first attempts to employ deep CNNs with a large amount data for face recognition. DeepFace adopts an AlexNet-like structure with similar input size, similar number of layers, large convolution kernel size, Rectified Linear Unit (ReLU)~\cite{nair2010rectified} and Dropout~\cite{srivastava2014dropout}. Comparatively, DeepIDs~\cite{sun2014deep1, sun2014deep2} use a smaller network structure. DeepID1 directly trains an 8-layer network on 10K classes with input image size $39\times 31\times 1$, and DeepID2 combines the verification loss and identification loss and changes the network structure with input image size $55\times 47\times 3$. VGGFace~\cite{parkhi2015deep} uses a bigger network structure of 19 layers with 2.6M training images. FaceNet~\cite{schroff2015facenet} uses the deep GoogleNet as the base network and adopts a triplet loss function. The development of residual network (ResNet)~\cite{he2016deep} enables the training of extremely deep networks possible. SphereFace~\cite{liu2017sphereface} makes use of a 64-layer ResNet with a newly proposed large angular margin loss. The Coco loss~\cite{liu2017rethinking} applies a 128-layer ResNet for face recognition. 

In DNN based face recognition, the selection of network structures often depends on the number and size of images as well as the computational power. There is no general agreement on which network structures should be used. Residual module and Google Inception module are the two most popular modules. In this work, we adopt the Google Inception\_ResNet\_V1~\cite{szegedy2017inception} model as our network structure due to its high efficiency and effectiveness.

\subsection{Loss Function}
\textbf{Softmax Loss.} Softmax loss~\cite{lecun1998gradient, krizhevsky2012imagenet} is a standard multi-class classification loss function in DNN. It projects an input feature, learned by deep neural network, into a probability distribution. In softmax loss, the predicted posterior probability for the $k$-th class is defined as follows:

\begin{equation}
 p_{k} = \frac{\exp(\mathbf{w}_{k}^T \mathbf{x} + b_{k})}{\sum_{l=1}^K{\exp(\mathbf{w}_{l}^T \mathbf{x} + b_{l})}}, 
\label{eq:softmax_eq1}
\end{equation}

\noindent where $\mathbf{x}$ is the feature vector in the last layer,
$\mathbf{w}_{l}$ and $b_{l}$ are classification vector and bias of the $l$-th class, $K$ is the total number of classes. 


With the probability distribution of the feature $\mathbf{x}_i$ of the $i$-th input sample, a cross-entropy loss can be calculated as follows:
\begin{equation}
 \mathcal{L}_{s} = \sum_{i}^{N}{-\log(p_{y_i})},
\label{eq:crossentropysoftmax_original_eq2}
\end{equation}

\noindent where $y_i$ is the label of the $i$-th sample and
$N$ is the total number of training samples.

The softmax loss has been used in DeepFace~\cite{taigman2014deepface} and DeepID~\cite{sun2014deep1} for face recognition. When the training of the DNN is done, the loss function will be removed and only the trained feature extractor is used in the deployment stage.





\noindent \textbf{Triplet Loss.} Triplet loss was firstly introduced in FaceNet~\cite{schroff2015facenet} to improve face recognition. It is originally proposed in the Large Margin Nearest Neighbor (LMNN) method~\cite{weinberger2009distance}.
Given a triplet $(\mathbf{x}_a,\mathbf{x}_p,\mathbf{x}_n)$ where $\mathbf{x}_a$ is the anchor sample, $\mathbf{x}_p$ denotes the positive sample, $\mathbf{x}_n$ represents the negative sample, the formulation of the triplet loss is defined as:

\begin{equation}
 \mathcal{L}_{t} = \sum_{i \in N} \left[ \left  \|\mathbf{x}_{a_i}-\mathbf{x}_{p_i}\right \|_{2}^{2}  - \left \| \mathbf{x}_{a_i}-\mathbf{x}_{n_i}\right \|_{2}^{2} + \epsilon \right ]_+, \nonumber
\label{eq:triplet}
\end{equation}

\noindent where $[\tau]_+$ denotes $\max(\tau,0)$, $\epsilon$ is a preset margin, and $N$ is the number of triplets. In triplet loss, the first term 
$\left  \|\mathbf{x}_{a_i}-\mathbf{x}_{p_i}\right \|_{2}^{2}$ is to pull samples from the same class together, and the second term $- \left \| \mathbf{x}_{a_i}-\mathbf{x}_{n_i}\right \|_{2}^{2}$ is to push samples from different classes away.


\noindent \textbf{Center Loss.} Center loss~\cite{wen2016discriminative} learns a center for each class by minimizing the distance between the sample features and their corresponding class centers. The definition of center loss is defined as follows:
\begin{equation}
 \mathcal{L}_{c} =  \sum_{i}^{N} \norm{{ \mathbf{x}_i - \mathbf{c}_{y_i}  } }_2^2, \nonumber
\label{eq:softmax2}
\end{equation}
\noindent where $\mathbf{x}_i$ is the $i$-th sample feature vector, $\mathbf{c}_{y_i}$ is the center vector of the $y_i$-th class, and $N$ is the number of samples. In practice, a weighted version of the softmax loss $\mathcal{L}_{s}$ and the above-mentioned center loss is usually used:
\begin{equation}\label{eq:12}
\begin{split}
\mathcal{L}_{sc} &= \mathcal{L}_{s} + \lambda \mathcal{L}_{c}, \nonumber
\end{split}
\end{equation}

\noindent where $\lambda$ is a balance parameter.


\noindent \textbf{SphereFace Loss.} Liu~\etal~\cite{liu2017sphereface} pointed out that the original softmax loss $\mathcal{L}_{s}$ in Eq.~\ref{eq:crossentropysoftmax_original_eq2} is rational only for close-set problems, but have disadvantages for face recognition, which is an open-set problem\footnote{In an open-set problem, unknown classes may occur in the test stage. In a close-set problem, all test classes are known in the training stage.}. They reformulated the original softmax loss into a modified softmax form:

\begin{equation}
\mathcal{L}_{ms} = \sum_{i}^{N} - \log ( \frac{\exp(\frac{\mathbf{w}_{y_i}^T\mathbf{x}_i}{\norm{\mathbf{w}_{y_i}}} )}{\sum_{k=1}^K{\exp(\frac{\mathbf{w}_{k}^T\mathbf{x}_i}{\norm{\mathbf{w}_{k}}} ) }}). \nonumber
\label{eq:softmax2}
\end{equation}

The bias term $b$ is removed and the weight $\mathbf{w}$ is normalized by its $L_2$ norm. The normalized classification vector $\frac{\mathbf{w}}{\norm{\mathbf{w}}}$ will lie on a hypersphere.
The modified softmax loss $\mathcal{L}_{ms}$ reduces the inconsistency of similarity measures in training and test stages brought by the bias term $b$ and the magnitude of classification vector $\mathbf{w}$.

The modified softmax loss $\mathcal{L}_{ms}$ can be rewritten into an angular version:

\begin{equation}
\mathcal{L}_{as} = \sum_{i}^{N} - \log ( \frac{\exp(\norm{\mathbf{x}_i} \cos({\theta}_{y_i, i}) )}{\sum_{k=1}^K{\exp(\norm{\mathbf{x}_i} \cos({\theta}_{k, i}))}} ),
\label{eq:angular_sphereface_eq3}
\end{equation}

\noindent which is called A-Softmax. Liu~\etal further introduced an angular margin to A-Softmax:


\begin{equation}
\mathcal{L}_{sphere} = \sum_{i}^{N} - \log ( \frac{\omega(\mathbf{x}_i, y_i)}{\omega(\mathbf{x}_i, y_i) + \psi(\mathbf{x}_i, y_i) }), 
\label{eq:angular_sphereface_mm_eq4}
\end{equation}

\noindent where

\begin{equation} 
\begin{split}
\omega(\mathbf{x}_i, y_i) = {\exp(\norm{\mathbf{x}_i} \cos(m{\theta}_{y_i, i}))}, \nonumber \\
\psi(\mathbf{x}_i, y_i) = \sum_{k\neq y_i}{\exp(\norm{\mathbf{x}_i} \cos({\theta}_{k, i}))}.  \nonumber
\end{split}
\end{equation}




This angular margin version of A-Softmax is named as SphereFace~\cite{liu2017sphereface}. The
SphereFace loss provides a clear understanding on how the classification vector $\mathbf{w}$ affects face recognition; however, it does not discuss the importance of the distribution of $\mathbf{x}$.

\noindent {\bf{Remarks.}} In addition to the above-mentioned loss functions, some recent works~\cite{ranjan2017l2, wang2017normface,liu2017rethinking} impose an $L_2$-constraint on the features so that the feature vectors are restricted to lie on a hypersphere. The $L_2$-constrained features could be written as follows:

\begin{equation}
 \mathbf{\hat x} = \alpha \frac{\mathbf{x}}{\norm{\mathbf{x}}},
\label{eq:angular_sphereface}
\end{equation}
\noindent where $\alpha$ is a preset or learnable parameter. 


The $L_2$-constraint operator normalizes the features to a fixed hypersphere so that each sample will contribute equally to the final cross-entropy loss function for network updating. However, one possible problem is that the loss function will become sensitive to the choice of $\alpha$. If $\alpha$ is small, the distribution of projected probabilities $p_k$ via softmax operator in Eq.~\ref{eq:softmax_eq1} will become very flat, reducing the discrimination of the cross-entropy loss function in Eq.~\ref{eq:crossentropysoftmax_original_eq2}. If $\alpha$ is large, the probabilities $p_k$ will vary much, reducing the stability of network training.


Our work is inspired by the SphereFace loss. However, we observe that in the training process of a DNN, the final convergence status is largely dependent on the distribution of $\mathbf{x}$, and the similarity matching in the test stage is basically determined by $\mathbf{x}$ as well. Therefore, different from SphereFace which focuses on formulating $\mathbf{w}$, we aim to formulate simultaneously $\mathbf{x}$ and $\mathbf{w}$. In SphereFace, the angular margin is an extremely important factor for the final performance of face recognition. 
In contrast, our model works consistently well on different face recognition tasks even without using any angular margin.



\section{Centralized Coordinate Learning}

We first present our formulation of the loss function, then introduce the centralized feature learning strategy. After that, we discuss the relation of our model to previous works. Finally, an adaptive angular margin is proposed to further improve the discriminative power of the learned features.

\subsection{Formulation of Loss Function}
\label{sec:3.1}
In a standard DNN for classification, an inner product operator is employed in the softmax operation for probability projection (refer to Eq.~\ref{eq:softmax_eq1}). The inner product is defined as:
\begin{equation}
z = \mathbf{w}_k^{T} \mathbf{x} + b_k, 
\label{eq:wxb_eq6}
\end{equation}
\noindent where $\mathbf{w}_k$ and $b_k$ are the classification vector and the bias for the $k$-th class,  and $\mathbf{x}$ is the output feature at the last layer of the neural network.

Both the features $\mathbf{x}$ and the classification vectors $\mathbf{w}$ are variables to be learned by the DNN. They are iteratively and alternatively updated in the training process. Given the optimization method and the hyper parameters, the convergence of the network mostly depends on the setting of $\mathbf{w}$ and the distribution of $\mathbf{x}$. Therefore, how to formulate the features $\mathbf{x}$ and the classification vectors $\mathbf{w}$ is crucial to the loss function design and consequently the learning outputs. An improper formulation of $\mathbf{w}$ may induce a classification gap between the training and test stages, while a less effective formulation of $\mathbf{x}$ may lead to less discriminative distribution of learned features. Researchers have proposed several schemes~\cite{ranjan2017l2, wang2017normface,liu2016large, liu2017rethinking, liu2017sphereface} to transform either $\mathbf{w}$ or $\mathbf{x}$ before performing the inner product. In this paper, we propose to transform both $\mathbf{w}$ and $\mathbf{x}$ for a more effective network training and face matching.


\begin{figure*}
	\centering
	\begin{subfigure}{0.35\textwidth} 
		\includegraphics[width=1.0\textwidth]{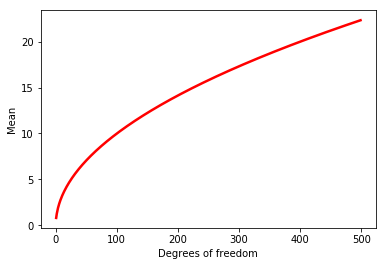}
		\caption{The mean of $r$.} 
	\end{subfigure}
	\vspace{1em} 
	\begin{subfigure}{0.36\textwidth} 
		\includegraphics[width=1.0\textwidth]{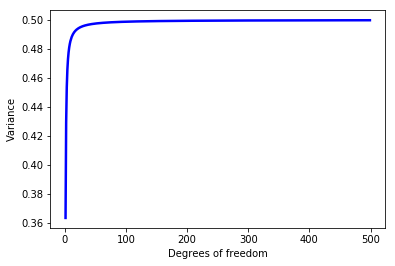}
		\caption{The variance of $r$.} 
	\end{subfigure}
	\caption{Mean and variance of variable $r$ (Eq.~\ref{eq:ourmain_eq12}) with respect to its degrees of freedom.} 
\end{figure*}

In open-set problems such as face recognition, the bias term $b$ may make the face similarity computation inconsistent in the training and test stages. As in~\cite{liu2016large, liu2017sphereface}, we disable this term in Eq.~\ref{eq:wxb_eq6}. We then introduce two transformations $\Psi(\cdot)$ and $\Phi(\cdot)$ on $\mathbf{w}$ and $\mathbf{x}$, respectively, resulting in the following inner product:
\begin{equation}
z = {{\Psi(\mathbf{w})}^{T}} \Phi(\mathbf{x}).
\label{eq:oureq1_eq7}
\end{equation}

\noindent The choices of $\Psi(\cdot)$ and $\Phi(\cdot)$ will largely influence the softmax probability $p_k$ of a sample, and thus impact the final loss function. Note that some recent works~\cite{ liu2016large, liu2017sphereface,wang2017normface, liu2017rethinking, ranjan2017l2} can be written as Eq.~\ref{eq:oureq1_eq7} with specific forms of $\Psi(\cdot)$ and $\Phi(\cdot)$.

In~\cite{liu2017sphereface}, Liu~\etal discussed the formulation of $\mathbf{w}$, and they let $\Psi(\mathbf{w}) = \frac{\mathbf{w}}{\norm{\mathbf{w}}}$. Such an $L_2$ normalization eliminates the influence of the varying magnitude of $\mathbf{w}$ on face matching in the test stage. Note that in the training stage of DNN, we compute $z ={\Psi(\mathbf{w})}^{T}\Phi(\mathbf{x})$ to update the network, but in the test stage we often employ the cosine similarity to match two face feature vectors $\Phi(\mathbf{x}_1)$ and $\Phi(\mathbf{x}_2)$:

\begin{equation}
s = \frac{{\Phi(\mathbf{x}_1)}^{T}}{\norm{\Phi(\mathbf{x}_1)}} \frac{\Phi(\mathbf{x}_2)}{\norm{\Phi(\mathbf{x}_2)}},
\label{eq:oureq1_eq8}
\end{equation}

\noindent where the trained $\mathbf{w}$ is not involved. Normalizing the classification vectors $\mathbf{w}$ in DNN training could avoid the case that some classification vectors with big magnitudes dominate the training of face features $\mathbf{x}$.


By setting $\Psi(\mathbf{w}) = \frac{\mathbf{w}}{\norm{\mathbf{w}}}$ in Eq.~\ref{eq:oureq1_eq7}, the softmax loss in Eq.~\ref{eq:crossentropysoftmax_original_eq2} can be rewritten as:
\begin{equation}
\mathcal{L}_{sf} = \sum_{i}^{N} - \log ( \frac{\exp(\frac{\mathbf{w}_{y_i}^T \Phi(\mathbf{x}_i)}{\norm{\mathbf{w}_{y_i}}} )}{\sum_{k=1}^K{\exp(\frac{\mathbf{w}_{k}^T\Phi(\mathbf{x}_i)}{\norm{\mathbf{w}_{k}}} ) }}).
\label{eq:oureq2_eq9}
\end{equation}

\noindent According to the definition of cosine similarity, Eq.~\ref{eq:oureq2_eq9} can be reformulated into an angular version as follows:
\begin{equation}
\mathcal{L}_{sf} = \sum_{i}^{N} - \log ( \frac{\exp(\norm{{\Phi(\mathbf{x}_i)}} \cos({\theta}_{y_i, i}) )}{\sum_{k=1}^K{\exp(\norm{{\Phi(\mathbf{x}_i)}} \cos({\theta}_{k, i}))}} ),
\label{eq:softmax_angle_eq10}
\end{equation}

\noindent where ${\theta}_{y_i, i}$ is the intersection angle between $\Phi(\mathbf{x}_i)$ and $\frac{\mathbf{w}}{\norm{\mathbf{w}}}$, and the range of ${\theta}_{y_i, i}$ is $[0, \pi]$. One can see that $\norm{{\Phi(\mathbf{x}_i)}}$ and $\cos({\theta}_{y_i, i})$ will impact the loss function in Eq.~\ref{eq:softmax_angle_eq10}, and both of them depend on the form of ${\Phi(\mathbf{x}_i)}$.


\subsection{Centralized Feature Learning}
\label{sec:3.2}
From the analysis in Section~\ref{sec:3.1}, we can see that the formulation of $\Phi(\mathbf{x})$ will be the key factor to the success of face feature learning. On one hand, it largely affects the training of DNN via loss function in Eq.~\ref{eq:softmax_angle_eq10}. If $\norm{{\Phi(\mathbf{x}_i)}}$ is small, the softmax projected probabilities of all samples $\Phi(\mathbf{x}_i)$ will become similar so that the loss function will be less discriminative. If $\norm{{\Phi(\mathbf{x}_i)}}$ is large, the probabilities may vary much and make the learning of DNN less stable. On the other hand, in the test stage, the cosine similarity of two face feature vectors, as given in Eq.~\ref{eq:oureq1_eq8}, is computed for face recognition. Ideally, $\Phi(\mathbf{x})$ is expected to distribute dispersedly across the whole coordinate space so that two face feature vectors from different subjects can be more separable with big angles. 

In this paper, we propose to centralize the face features to the origin of the space during the learning process. Specifically, for each dimension $j$ of the feature vector $\mathbf{x}$, we define $\Phi(\mathbf{x}(j))$ as:

\begin{equation}
\Phi(\mathbf{x}(j))  =  \frac{\mathbf{x}(j) - \mathbf{o}(j) }{\boldsymbol{\sigma}(j)},
\label{eq:ourmain_eq11}
\end{equation}

\noindent where $\mathbf{o} = \mathop{\mathbb{E}}[\mathbf{x}]$ is the mean vector of $\mathbf{x}$, and $\boldsymbol{\sigma}(j)$ is the standard deviation of $\mathbf{x}(j)$. Clearly, the transform $\Phi(\cdot)$ will centralize each dimension of $\mathbf{x}$ to the origin so that the features $\Phi(\mathbf{x})$ will span across all quadrants of the coordinate space. Meanwhile, each dimension of $\Phi(\mathbf{x})$ will have the same unit variance so that each dimension will contribute equally to the discrimination of faces instead of using only several strong dimensions for face recognition. Actually, several recent works~\cite{ranjan2017l2, wang2017normface, liu2017rethinking} have been proposed to normalize $\mathbf{x}$ into a hypersphere as $\mathbf{\hat x} = \alpha \frac{\mathbf{x}}{\norm{\mathbf{x}}}$. However, the normalization operator will not change the quadrant of the feature $\mathbf{x}$ and the selection of parameter $\alpha$ is not a trivial work.

Let's then analyze the $L_2$ norm of $\Phi(\mathbf{x})$, i.e., $\norm{\Phi(\mathbf{x})}$. With the transform in Eq.~\ref{eq:ourmain_eq11}, it is reasonable to assume that $\Phi(\mathbf{x}(j)), j=1,2,...,D$, are i.i.d. variables and each variable follows a standard Gaussian distribution $\mathcal{N}(0, 1)$. Then the $L_2$ norm of $\Phi(\mathbf{x})$, defined as
\begin{equation}
r = \norm{\Phi(\mathbf{x})} = \sqrt{\sum_{j=1}^D {(\Phi(\mathbf{x}(j)))}^2},
\label{eq:ourmain_eq12}
\end{equation}

\noindent will follow the $\chi$-distribution with $D$ degrees of freedom~\cite{abramowitz1964handbook}. The mean and variance of $r$, denoted by ${\mu}_r$ and ${{\sigma}^2_r}$, are 

\begin{equation}
{\mu}_r  = \mathop{\mathbb{E}}[r] = \sqrt{2} \frac{{\Gamma} ((D+1)/2)}{{\Gamma} {(D/2)}},
\label{eq:ourmain_eq13}
\end{equation}

\begin{equation}
{{\sigma}^2_r}  = \mathop{\mathbb{Var}}[r] = D - {{\mu}^2_r},
\label{eq:ourmain_eq14}
\end{equation}

\noindent where ${\Gamma}(\cdot)$ is a gamma function.
Both ${\mu}_r$ and ${{\sigma}^2_r}$ are functions of $D$. In Figs. 1(a) and 1(b), we plot the curves of ${\mu}_r$ and ${{\sigma}^2_r}$ with respect to the degrees of freedom $D$. 

As can be seen from Fig. 1(a), ${\mu}_r$ increases with the increase of $D$. Actually, ${\mu}_r$ is very close to $\sqrt{D}$. For example, when $D$ = 400, $r$ = 19.987. As we discussed before, a small $\norm{\Phi(\mathbf{x})}$ will make the softmax projected probabilities of face samples similar to each other so that the discriminability will be reduced, while a large $\norm{\Phi(\mathbf{x})}$ will reduce the stability of network training. Based on our experience, setting $D$ between 350 and 400 will be a rational choice. (In our implementation, $D$ is set to 374.)  From Fig. 1(b), we can see that ${{\sigma}^2_r}$ is around 0.5 when $D$ is larger than 20. It is more than one order smaller than ${\mu}_r$. This is a very desirable property because it implies that for most of the face samples, their corresponding $\norm{\Phi(\mathbf{x})}$ values will not vary much. Therefore, each sample will approximately contribute equally to the final cross-entropy loss function for network updating.



\begin{figure*}[h]
\includegraphics[scale=0.45]{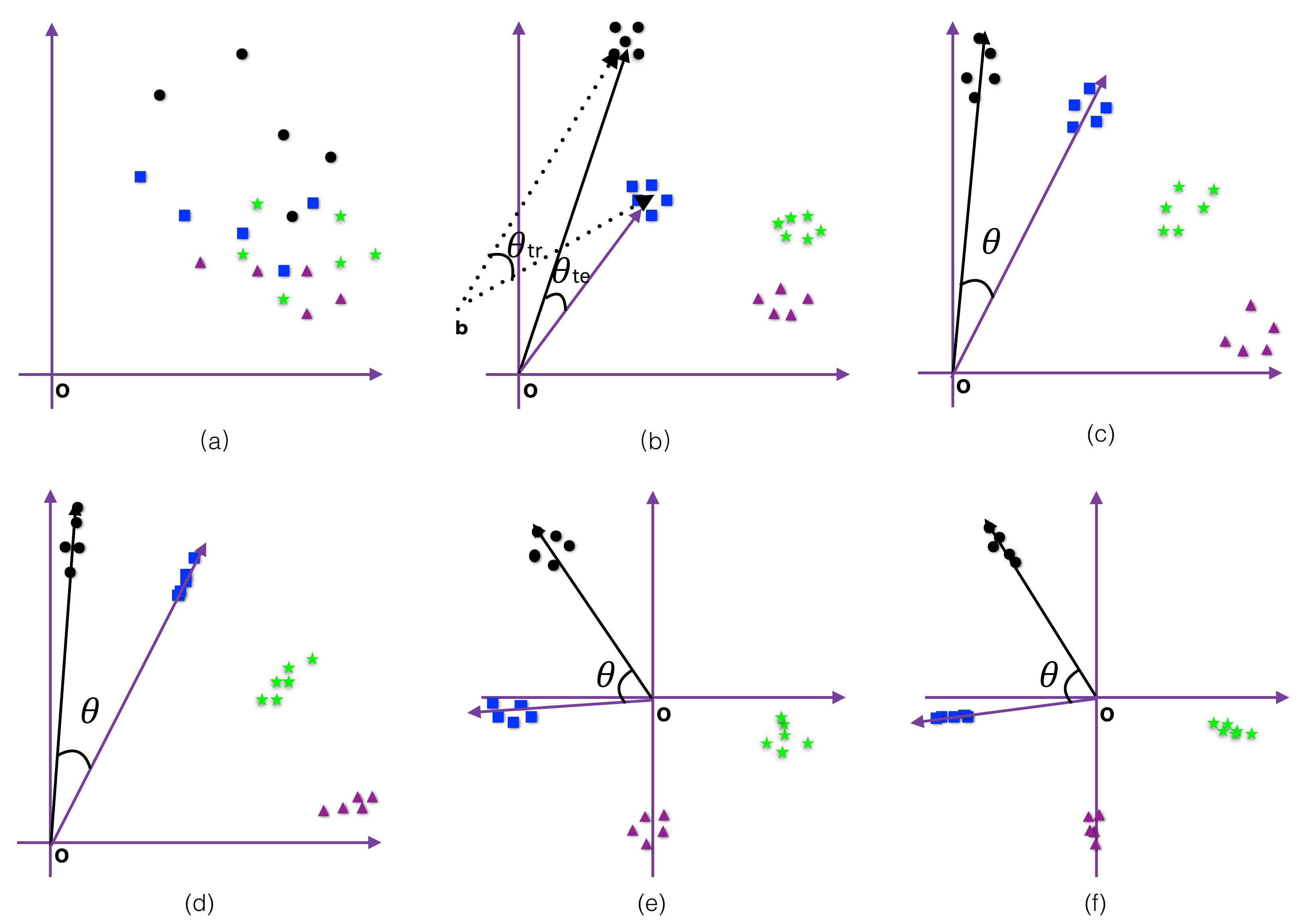}
\centering
\caption{Illustration of the effects of different loss functions. The ``black dots'', ``blue squares'', ``green stars'' and ``pink triangles'' represent samples of four face classes. (a). Original data distribution. (b). Converged face features by using the original softmax loss. (c). Converged face features by using the A-Softmax loss. (d) Converged face features by using the SphereFace loss. (e) Converged face features by using the CCL loss. (f). Converged face features by using the CCL loss with AAM.}
\label{fig:visualaffect}
\end{figure*}

Having analyzed the properties of the proposed transform $\Phi(\mathbf{x})$, now let us discuss how to learn the origin vector $\mathbf{o}$ and the standard deviation $\boldsymbol{\sigma}$ in Eq.~\ref{eq:ourmain_eq11}. Given two face feature vectors $\mathbf{x}_1$ and $\mathbf{x}_2$, their cosine similarity by taking $\mathbf{o}$ as the coordinate origin is:

\begin{equation}
s_\mathbf{o}(\mathbf{x}_1, \mathbf{x}_2) = 
\frac{(\mathbf{x}_1-\mathbf{o})^{T}{(\mathbf{x}_2-\mathbf{o})}}   {\norm{\mathbf{x}_1-\mathbf{o}} \norm{\mathbf{x}_2-\mathbf{o}}}
\label{eq:cosinesimilarity}
\end{equation}

The similarity between  $\mathbf{x}_1$ and $\mathbf{x}_2$ is sensitive to the origin $\mathbf{o}$. For instance, let $\mathbf{x}_1$ be [0.1, 0.1, 0.2] and $\mathbf{x}_2$ be [0.2, 0.2, 0.1], if the origin is [0, 0, 0], the intersection angle between $\mathbf{x}_1$ and $\mathbf{x}_2$ is ${35.3}^\circ$ and $s_\mathbf{o}(\mathbf{x}_1, \mathbf{x}_2) = 0.816$. If the origin moves a little to [-0.01, -0.01, -0.01], then the intersection angle becomes ${33.1}^\circ$ and $s_\mathbf{o}(\mathbf{x}_1, \mathbf{x}_2) = 0.837$.
Thus, in each update during DNN training, we should not change $\mathbf{o}$ and $\boldsymbol{\sigma}$ too much. We use a moving average strategy with a large decay factor to update $\mathbf{o}$ and $\boldsymbol{\sigma}$:


\begin{equation}
  \begin{cases}
    \mathbf{o}_{new} & = \mathcal{\rho}\cdot \mathbf{o}_{old} + (1-\mathcal{\rho})\cdot \mathbf{o}_{b}, \\
    \boldsymbol{\sigma}_{new} & = \mathcal{\rho}\cdot \boldsymbol{\sigma}_{old} + (1-\mathcal{\rho})\cdot \boldsymbol{\sigma}_{b},
  \end{cases}
\end{equation}

 

\noindent where $\mathcal{\rho}$ is the decay factor, and $\mathbf{o}_{b}$ and $\boldsymbol{\sigma}_{b}$ are the mean and standard deviation vectors generated by the current mini-batch. In our implementation, $\mathcal{\rho}$
is set as 0.995 to ensure that $\mathbf{o}$ and $\boldsymbol{\sigma}$ do not change too much in each iteration.



\subsection{Relation to Previous Works}
\label{sec:3.3}
Overall, the proposed formulation of centralized coordinate learning (CCL) can be written as:
\begin{equation}
z = \frac{\mathbf{w^T}}{\norm{\mathbf{w}}}\Phi(\mathbf{x}),
\label{eq:ourmain_3333}
\end{equation}

\noindent where $\Phi(\mathbf{x})$ is defined in Eq.~\ref{eq:ourmain_eq11}. In CCL, the normalization on classification vector $\mathbf{w} = \frac{\mathbf{w}}{\norm{\mathbf{w}}}$ is the same as SphereFace~\cite{liu2017sphereface}, which can be considered as a simplified version of {\it{Weight Normalization}} (WN)~\cite{salimans2016weight}. However, the normalization is only applied to the classification layer instead of each layer of the network as in WN.

As for the feature transformation $\Phi(\mathbf{x})$, it shares a similar form to the widely used {\it{Batch Normalization}} (BN)~\cite{ioffe2015batch} if we add scaling and shifting terms to it:

\begin{equation}
\Phi(\mathbf{x}(j))  =  \boldsymbol{\gamma}(j) \frac{\mathbf{x}(j) - \mathbf{o}(j) }{\boldsymbol{\sigma}(j)} + \boldsymbol{\beta}(j).
\label{eq:ourmain_18}
\end{equation}

\noindent However, BN is applied to each layer after the convolution operation, while CCL is only applied to the last classification layer before softmax operation. Furthermore, the parameters $\boldsymbol{\gamma}(j)$ and $\boldsymbol{\beta}(j)$ in BN may destroy the advantages of $\Phi(\mathbf{x})$ analyzed in Section~\ref{sec:3.2}. We will make more discussions on this Section~\ref{sec:4.2}.

\subsection{Adaptive Angular Margin}
\label{sec:3.4}
Angular margin was firstly introduced in L-Softmax~\cite{liu2016large} and SphereFace~\cite{liu2017sphereface} to make the classification boundary more compact. It has proved to be an effective way to further improve face recognition performance. However, the angular margin function introduced in~\cite{liu2016large, liu2017sphereface} is difficult to train and is sensitive to parameters. To ease this issue, we propose a simple adaptive angular margin (AAM) function as follows:

\begin{equation}
\mathcal{L}_{AAM} = \sum_{i}^{N} - \log (p^{AAM}_{y_i}),
\label{eq:adaptiveam_eq19}
\end{equation}

\noindent where \\


\begin{equation}
{p^{AAM}_{y_i}} =  \frac{\exp(\norm{{\Phi(\mathbf{x}_i)}} \cos(\eta{\theta}_{y_i, i}) )}{\exp(\norm{{\Phi(\mathbf{x}_i)}} \cos(\eta{\theta}_{y_i, i}) + \sum_{k\neq y_i} {\exp(\norm{{\Phi(\mathbf{x}_i)}} \cos({\theta}_{k, i}))}},
\label{eq:adaptiveam111_eq20}
\end{equation}


\noindent where $\eta$ is an adaptive parameter and it is set based on the value of ${\theta}_{y_i, i}$:
\begin{equation}
 \eta = \left\{ \begin{array}{ll}
   1, & \pi/3<{\theta}_{y_i, i}<=\pi;  \\
   \frac{\pi/3}{{\theta}_{y_i, i}}, & \pi/30<{\theta}_{y_i, i}<=\pi/3;  \\
   10, & {\theta}_{y_i, i}<= \pi/30.  
 \end{array}  \right. 
\label{eq:adaptiveam111_eq21}
\end{equation}



As shown in Eq.~\ref{eq:adaptiveam111_eq21}, we partition the range of ${\theta}_{y_i, i}$ into three intervals, based on which $\eta$ is set. When $\pi/3<{\theta}_{y_i, i}<=\pi$, the angle is big enough to provide enough gradient information for back-propagation, thus, adding a stronger angular margin is not necessary. For example, if ${\theta}_{y_i, i} = \frac{2\pi}{3}$ and $\eta$ is set to 2, then the quadrant will jump from the second quadrant ($\frac{2\pi}{3}$) to the third quadrant ($\frac{4\pi}{3}$), which will bring in wrong gradient information. Therefore, when $\pi/3<{\theta}_{y_i, i}<=\pi$, we set $\eta = 1$. When $\pi/30<{\theta}_{y_i, i}<=\pi/3$, the angle is relatively small and it is necessary to introduce additional angular margin, and we set $\eta = \frac{\pi/3}{{\theta}_{y_i, i}}$, i.e., the smaller the angle ${\theta}_{y_i, i}$, the more angular margin we introduce. When ${\theta}_{y_i, i}<= \pi/30$, we fix the parameter $\eta=10$ since a too big $\eta$  may make the back-propagation vibrate too much. 

The AAM loss defined above can pull the face feature vectors from the same subjects more compactly distributed, but using it alone to train the DNN may not be stable. In practice, we leverage a weighted version of the softmax loss in Eq.~\ref{eq:softmax_angle_eq10} and the AAM loss in Eq.~\ref{eq:adaptiveam_eq19} for training:

\begin{equation}
\mathcal{L} = \frac{\lambda \mathcal{L}_{sf} + \mathcal{L}_{AAM} }{\lambda + 1.0},
\label{eq:adaptiveam_21}
\end{equation}


\noindent where $\lambda$ is a constant to balance $\mathcal{L}_{sf}$ and $\mathcal{L}_{AAM}$. In our implementation, we empirically set $\lambda = 3$.




In Fig. 2, we illustrate the differences of different loss functions on the convergence of face features, including the original softmax loss $\mathcal{L}_{s}$ in Eq.~\ref{eq:crossentropysoftmax_original_eq2}, the A-Softmax loss in Eq.~\ref{eq:angular_sphereface_eq3}, the SphereFace loss in Eq.~\ref{eq:angular_sphereface_mm_eq4}, the CCL loss $\mathcal{L}_{sf}$ in Eq.~\ref{eq:softmax_angle_eq10}, and the loss of CCL with AAM in Eq.~\ref{eq:adaptiveam_eq19}. From Fig. 2, we have the following comments:

\begin{itemize}[leftmargin=0.5cm]
\item As shown in Fig. 2(b), the original softmax loss $\mathcal{L}_{s}$ may make the cosine similarity of face vectors inconsistent in the training and test stages, due to the bias term $b$ and the different magnitudes of $\mathbf{w}$ (i.e., the lengths of different class centers can be very different). The angle ${\theta}_{tr}$ between two classes (denoted by the black circle and blue square, respectively) in the training stage may be different from the angle ${\theta}_{te}$ in the test stage.
    
\item The A-Softmax loss can eliminate this similarity inconsistency in training and test stages by removing the bias term $b$ and normalizing the magnitude of classification vectors $\mathbf{w}$. As shown in Fig. 2(c), the centers of different classes lie nearly in a unit hypersphere. The SphereFace loss, by introducing a large angular margin, can enforce the face samples from the same subject to get closer to each other, as shown in Fig. 3(d). This will enhance the discrimination of face features.


\item Suppose that the original face features lie in the first quadrant, as shown in Fig. 2(a), the softmax loss, A-Softmax loss and SphereFace loss will pull each class of face images to its center, while push different classes away from each other. However, most of the face features will still lie in the first quadrant. By using the proposed CCL based loss, the learned face features will be centralized to a common origin so that they will span across all the four quadrants. Intuitively, such a disperse distribution of face features tends to make the neighboring classes have larger angles, improving their separability, as illustrated in Fig. 2(e). With the proposed AAM, the intra-class variations can be further reduced, as shown in Fig. 2(f). In this way, the AAM can further improve the discrimination capability of face features.
\end{itemize}

\begin{table*}[t]
\centering
\scriptsize
\caption{Inception\_ResNet\_V1 network structure used in this paper.}
\begin{tabular}{ |c|c|c|c|c|c|c|c| } 
 \hline
 Layer & size-in & size-out & kernel &stride,padding & params   &ReLU\_fn & scale\\ [0.06cm]
 \hline
 Conv\_BN\_ReLU   & $160\times 160 \times 3$   & $79\times 79 \times 32$ & $3\times 3\times 3$  & 2, 0  & $32\times 3\times 3\times 3$ &  True  & - \\[0.06cm]
 Conv\_BN\_ReLU   & $79\times 79 \times 32$    & $77\times 77 \times 48$ & $3\times 3\times 32$ & 1, 0  & $48\times 3\times 3\times 32$ &  True  & - \\[0.06cm]
 Conv\_BN\_ReLU   & $77\times 77 \times 48$    & $77\times 77 \times 64$ & $3\times 3\times 48$ & 1, 1  & $64\times 3\times 3\times 48$ &  True  & - \\[0.06cm]
 MaxPool2D        & $77\times 77 \times 64$    & $38\times 38 \times 64$ & $3\times 3$          & 2, -  & 0  &  True  & - \\[0.03cm]
 Conv\_BN\_ReLU   & $38\times 38 \times 64$   & $38\times 38 \times 80$ & $1\times 1\times 64$  & 2, 0  & $80\times 1\times 1\times 64$ &  True  & - \\[0.06cm]
 Conv\_BN\_ReLU   & $38\times 38 \times 80$   & $36\times 36 \times 192$ & $3\times 3\times 80$ & 1, 0  & $192\times 3\times 3\times 80$  &  True  & - \\[0.06cm]
 Conv\_BN\_ReLU   & $36\times 36 \times 192$   & $17\times 17 \times 256$ & $3\times 3\times 192$  & 2, 0  & $256\times 3\times 3\times 192$   &  True  & - \\[0.06cm]
 $5\times$ Inception\_A   & $17\times 17 \times 256$   & $17\times 17 \times 256$ & Inception\_A  & -  & -   &  True  & 0.17 \\[0.06cm]
 Reduction\_A   & $17\times 17 \times 256$   & $8\times 8 \times 896$ & Reduction\_A  & -  & - &  True  & - \\[0.06cm]
 $10\times$ Inception\_B   & $8\times 8 \times 896$   & $8\times 8 \times 896$ & Inception\_B  & -  & -  &  True  & 0.1 \\[0.06cm]
 Reduction\_B   & $8\times 8 \times 896$   & $3\times 3 \times 1792$ & Reduction\_B  & -  & - &  True  & - \\[0.06cm]
 $5\times$ Inception\_C   & $3\times 3 \times 1792$   & $3\times 3 \times 1792$ & Inception\_C  & -  & - &  True  & 0.2 \\[0.06cm]
 Inception\_C   & $3\times 3 \times 1792$   & $3\times 3 \times 1792$ & Inception\_C  & -  & - &  False  & 1.0 \\[0.06cm]
 AvgPool2d   & $3\times 3 \times 1792$   & $1\times 1 \times 1792$ & $3\times 3$  & -  & 0 &  -  & - \\[0.06cm]
 \hline
\end{tabular}
\label{tab:network_structure}
\end{table*}




\section{Experiments}
We evaluate our method on six face datasets: Labeled Face in the Wild (LFW)~\cite{huang2008labeled}, Cross-Age Celebrity Dataset (CACD)~\cite{chen2014cross}, Cross-Age LFW (CALFW)~\cite{zheng2017cross}, Similar-Looking LFW (SLLFW)~\cite{deng2017fine}, YouTube Face (YTF)~\cite{wolf2011face} and MegaFace~\cite{kemelmacher2016megaface}.
On each dataset, we compare our method with the state-of-the-art results reported in literature.


\subsection{Experimental Details}
\label{sec:4.1}

{\bf{Training data.}} In this paper, we use only the CASIA WebFace dataset~\cite{yi2014learning} to train our CCL model. CASIA WebFace consists of 494,414 images of 10,575 subjects and it is widely used to train DNNs. In the original dataset, there are some annotation errors. A corrected version of CASIA Webface was later released, which has 455,594 images of 10,575 subjects. Compared to some previous works, such as DeepFace~\cite{taigman2014deepface} (4M), VGGFace~\cite{parkhi2015deep} (2M),  FaceNet~\cite{szegedy2015going} (200M) and Coco loss~\cite{liu2017rethinking} (half MS-Celebrity), the scale of CASIA WebFace (around 0.46M) is small. In addition, the number of face images in each class is uneven. Some classes have several hundreds of images, while some classes have only around 10 images. Regardless of the small scale and uneven distribution of face images in the CASIA WebFace dataset, it is a good platform to evaluate the effectiveness of a face learning algorithm without the influence of large-scale training data.

\noindent {\bf{Face Detection and Face Alignment.}} Face detection~\cite{zhang2016joint, yang2016wider, zhang2017s, hu2017finding} is an important step for the following face recognition and analysis. Following previous works~\cite{wen2016discriminative, liu2016large, liu2017sphereface}, in this paper we use MTCNN~\cite{zhang2016joint} for face detection in both training and test stages. In addition to providing the location of face, MTCNN also provides the positions of five face landmarks (the center of left eye, the center of right eye, nose, left mouth corner and right mouth corner). After the original image is inputted to the MTCNN face detector, if no face is detected in the image, we upsample the image by a factor of 2 and apply face detector again. If still no face is detected, a $182\times 182$ region is cropped from the center of the image as the detection output. 

Commonly used face alignment is to compute an affine transformation between the three landmarks detected by MTCNN and the three corresponding landmarks provided by a pre-defined face template such as OpenCV dlib library. However, simple affine transformation with three points correspondence may twist the profile faces. In this paper, we align the eyes to a horizontal line and make the center of three landmarks (the nose, the left mouth corner and the right mouth corner) in the middle of the image. A $182\times 182$ image is cropped from the center of the aligned image. (Note that all face images in figures shown in this paper are cropped images with size $182\times 182$.)

\noindent {\bf{Preprocessing and Data Augmentation.}} Each pixel of the cropped face image is normalized to range $[-1, 1]$ by first subtracting 127.5 and then dividing by 127.5. In the training stage, an online data augmentation strategy is used. For each image, a sub-image of size $160\times 160$ is randomly cropped from the pre-aligned $182\times 182$ image. These face images are randomly horizontally flipped before being inputted to the network.


\noindent {\bf{Network Structure.}} 
The proposed CCL method is implemented in PyTorch~\cite{paszke2017pytorch}. The Inception-ResNet network with a similar structure to the open FaceNet\footnote{https://github.com/davidsandberg/facenet} implementation is employed as the base network. Detailed information of our base network is listed in Table~\ref{tab:network_structure}, where ``scale'' refers to the scale factor used in residual block as $\mathbf{x} = \mathbf{x} +$scale $* \mathbf{f}(\mathbf{x})$. ReLU\_fn denotes where ReLU is applied in the last operator. 
For each image, we input itself and its horizontally flipped image into the based network, and obtain two 1,792-d feature vectors $\mathbf{f}$ and $\mathbf{f_{flip}}$ after the AvgPool2d operator as shown in Table~\ref{tab:network_structure}. We concatenate the two features using two different concatenation forms as $[\mathbf{f}, \mathbf{f_{flip}}]$ and $[\mathbf{f_{flip}}, \mathbf{f}]$ to obtain two 3,584-d features. These two features are embedded into two individual 374-d features with the same embedding weight matrices. Finally, CCL operator is applied to both of them.
In the training stage, the two features will lead to two individual loss values. 


\noindent {\bf{Learning Strategy.}}
We set the batch size as 128. The learning rate starts from 0.1, and then is divided by 10 at the 80K-th and 110K-th iterations. The total number of iterations is 150K. Weight decay is set to 0.0002. In all convolution layers, the bias term is disabled. It takes around 36 hours to finish 150K iterations on a Titan X Pascal GPU.


\noindent {\bf{Test settings.}} In the test stage, we input the $160\times 160$ image cropped from the center of the pre-aligned image and its corresponding flipped image to the trained network, and generate two 374-d feature vectors. These two feature vectors are averaged and then normalized into a unit-length vector. The similarity score between two face images is computed by the cosine distance. Threshold comparison is used in face verification.






\subsection{Exploratory Evaluation}
\label{sec:4.2}
\subsubsection{Evaluation of Centralized Coordinate Learning}
\label{sec:4.2.1}
To better illustrate the effectiveness of the proposed CCL strategy, here we conduct experiments to evaluate the performance of original linear embedding (LE) (i.e., only full-connected layer is used before softmax operation), LE with BN, LE with BN but without $\boldsymbol{\beta}$, and LE with CCL (CCL in short). The benchmark LFW database is used for the comparison experiments. The detailed information about LFW can be found in Section~\ref{sec:4.3}.


\begin{table}[h]
\centering
\small
\caption{Accuracy (\%)  on the LFW dataset.}
\begin{tabular}{ |c|c| } 
 \hline
 Method & LFW  \\ [0.08cm]
 \hline
 LE                                 & 98.367           \\ [0.08cm]
 LE with BN                         & 99.133           \\ [0.08cm]
 LE with BN (no $\boldsymbol{\beta}$)   & 99.217           \\ [0.08cm]
 CCL                                & 99.467      \\ [0.08cm]
 \hline
\end{tabular}
\label{tab:comparison11_1}
\end{table}


The experimental results are shown in Table~\ref{tab:comparison11_1}. One can see that CCL significantly outperforms LE, improving its accuracy from 98.367\% to 99.467\%. We can also see that CCL performs better than LE with BN. We observe that LE with BN (with $\boldsymbol{\gamma}$ but without $\boldsymbol{\beta}$) outperforms LE with original BN (with both $\boldsymbol{\gamma}$ and $\boldsymbol{\beta}$), while both of them are behind CCL. We believe that the parameters  $\boldsymbol{\gamma}$ and $\boldsymbol{\beta}$ in BN will bring certain negative effects on the distribution of features $\mathbf{x}$, which thus affect the final performance of instance-level face recognition problem.

\subsubsection{Evaluation of Adaptive Angular Margin}
\label{sec:4.2.2}
We then evaluate the effectiveness of the proposed AAM (refer to Eq.~\ref{eq:adaptiveam111_eq20}) by learning CCL models with and without AAM loss. We compare our methods with SphereFace with and without angular margins. The experimental results on LFW database are listed in Table~\ref{tab:comparison22}.


\begin{table}[h]
\centering
\small
\caption{Accuracy (\%) of CCL with and without adaptive angular margin on LFW dataset.}
\begin{tabular}{ |c|c| } 
 \hline
 Method & Accuracy \\
 \hline
 SphereFace without angular margin & 97.88  \\ [0.08cm]
 SphereFace with angular margin ($m=1$) & 97.90  \\ [0.08cm]
 SphereFace with angular margin ($m=2$) & 98.40  \\ [0.08cm]
 SphereFace with angular margin ($m=3$) & 99.25  \\ [0.08cm]
 SphereFace with angular margin ($m=4$) & 99.42 \\ [0.08cm]
 \hline
 CCL  & 99.467 \\[0.08cm]
 CCL with AAM & 99.583 \\[0.08cm]
 \hline
\end{tabular}
\label{tab:comparison22}
\end{table}

From Table~\ref{tab:comparison22}, it can be observed that the angular margin plays an important role in SphereFace. Without angular margin, SphereFace only has an accuracy of 97.88\%. With angular margin, the performance of SphereFace can be improved to 99.42\% ($m=4$). In comparison, our method, even without using angular margin, outperforms the best results of SphereFace with strong angular margin. AAM can further improve the performance of CCL on LFW. The difficulties of training DNN with angular margin will increase with the number of output classes because angular margin enforces harder conditions for the loss function. We also found that enforcing a strong angular margin sometimes makes the network fail to converge. Therefore, how to apply angular margin is still a worth pondering problem.

\begin{figure}[h]
\includegraphics[scale=0.16]{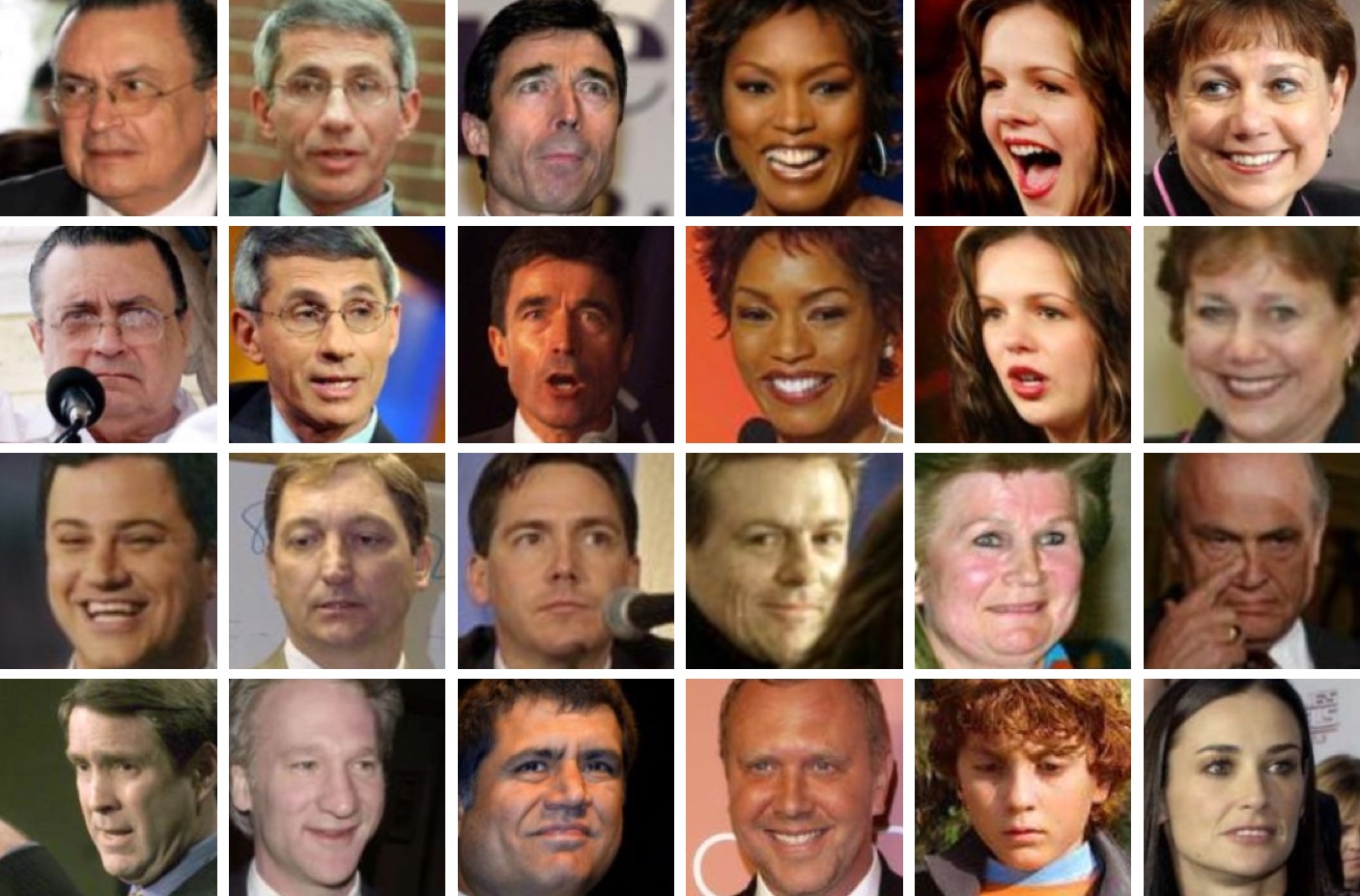}
\centering
\caption{Sample images from the LFW dataset. The first two rows show six positive pairs, and the last two rows show six negative pairs.}
\label{fig:LFWimage}
\end{figure}


\subsection{Experiments on Six Benchmarks}
\label{sec:4.3}
{\bf{Experiments on LFW.}} 
The LFW dataset~\cite{huang2008labeled} consists of 13,233 web-collected images from 5,749 different identities. Only 85 persons have more than 15 images, while 4,069 persons have only one image. There are 6,000 face pairs, including 3,000 positive pairs and 3,000 negative pairs. The 6,000 face pairs are divided into ten subsets, each having 300 positive pairs and 300 negative pairs. Note that the negative pairs are selected randomly. The images have different kinds of variations in pose, expression and illuminations. Some sample image pairs from LFW are shown in Fig.~\ref{fig:LFWimage}. 

Following the standard protocol of unrestricted with labeled outside data, we test CCL on 6,000 face pairs in comparison with the state-of-the-art methods. 
The experimental results are reported in Table~\ref{tab:mylfwtable}, from which we have the following observations:

\begin{itemize}
   \item The proposed CCL with AAM, trained on only the CASIA dataset with 0.46M samples, outperforms most state-of-the-art models, including those trained on much larger scale of data or the ensemble of multiple models. Its accuracy is only slightly lower than FaceNet~\cite{schroff2015facenet} and Coco Loss~\cite{liu2017rethinking}, which are trained with 200M samples and half MS-Cele data (about 2M)~\cite{guo2016ms}, respectively. 
   \item Compared with those models trained on the same CASIA dataset, including Center loss, L-Softmax, $L_2$ NormFace, and SphereFace, the proposed CCL models achieve the best accuracy, no matter the AAM loss is used or not. 
\end{itemize}

\begin{table}
\centering
\small
\caption{Accuracy (\%) on the LFW dataset. (``*'' denotes the ensemble of 25 models.)}
\begin{tabular}{ |c|c|c| } 
 \hline
 Method & Training Images & Accuracy \\
 \hline
DeepID2+~\cite{sun2015deeply}                &0.3M  &98.70 \\[0.08cm]
DeepID2+ (25)~\cite{sun2015deeply}           &0.3M*  &99.47 \\[0.08cm]
DeepFace~\cite{taigman2014deepface}          &4M  &97.35 \\[0.08cm]
Center Loss~\cite{wen2016discriminative}     &0.7M  &99.28 \\[0.08cm]
UP loss~\cite{guo2017one}                    &1.2M & 98.88 \\[0.08cm]
Marginal Loss~\cite{deng2017marginal}        &4M  &98.95 \\[0.08cm]
Noisy Softmax~\cite{chen2017noisy}                        &WebFace+  &99.18 \\[0.08cm]
Range Loss\cite{zhang2016range}              &1.5M  &99.52 \\[0.08cm]
FaceNet~\cite{schroff2015facenet}            &200M  &\bf{99.65} \\[0.08cm]
Coco Loss~\cite{liu2017rethinking}            &half MS-Cele~\cite{guo2016ms}  &\bf{99.86} \\[0.08cm]
\hline

Softmax Loss                           &CASIA  &97.88 \\[0.08cm]
Center Loss                            &CASIA  &99.05 \\[0.08cm]
Marginal Loss                          &CASIA  &98.95 \\[0.08cm]
$L_2$ NormFace~\cite{wang2017normface} &CASIA  &99.20 \\[0.08cm]
L-Softmax~\cite{liu2016large}          &CASIA  &99.10 \\[0.08cm]
ReST~\cite{Wu_2017_ICCV}               &CASIA  &99.05 \\[0.08cm]
SphereFace~\cite{liu2017sphereface}    &CASIA  &\bf{99.42} \\[0.08cm]

\hline
 CCL    &CASIA (0.46M) & \bf{99.467} \\ [0.08cm]
 CCL with AAM   & CASIA (0.46M)  & \bf{99.583} \\[0.08cm]
 \hline
\end{tabular}
\label{tab:mylfwtable}
\end{table}

\begin{figure}[t]
\includegraphics[scale=0.16]{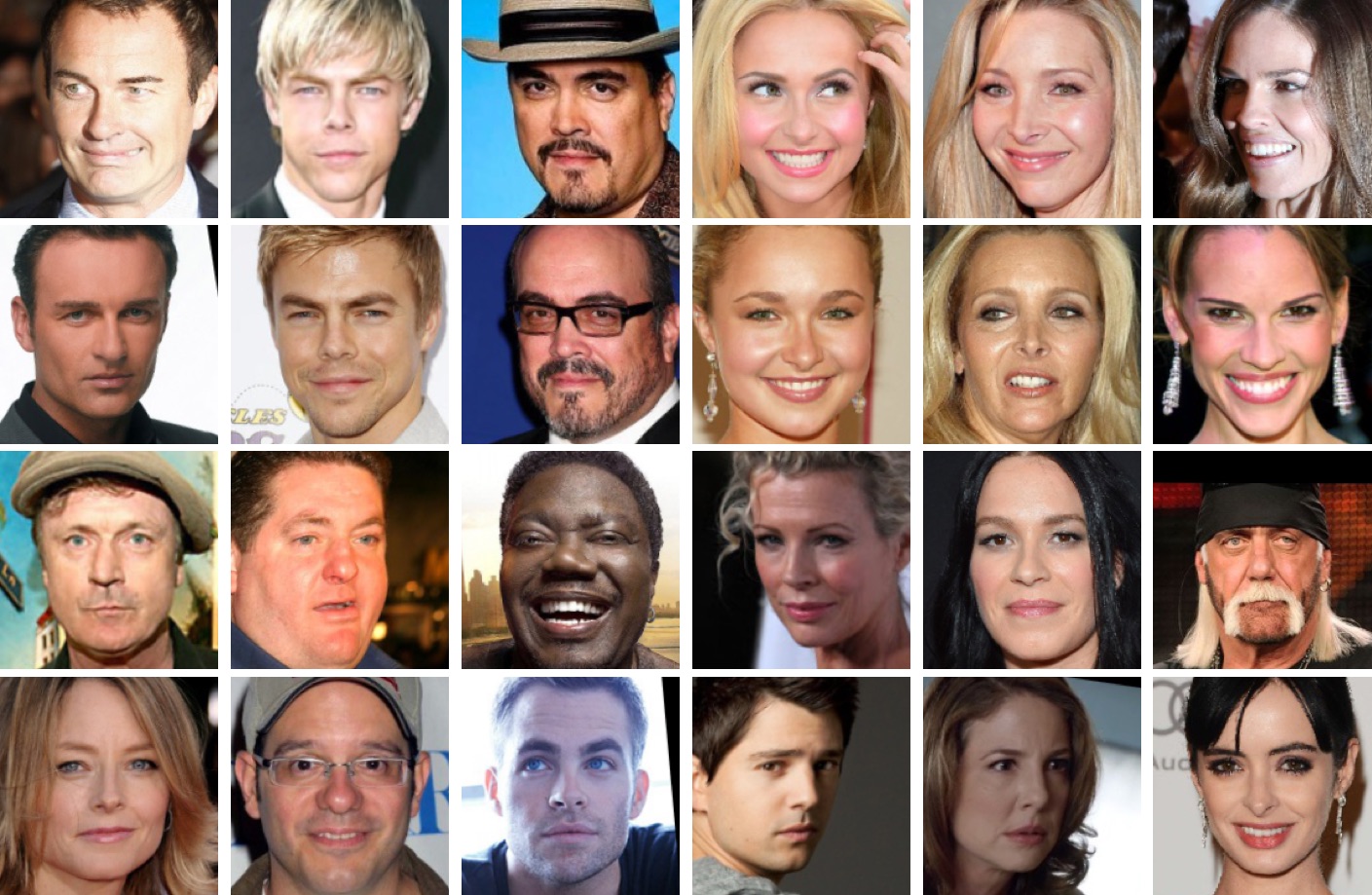}
\centering
\caption{Sample images from the CACD dataset. The first two rows show six positive pairs, and the last two rows show six negative pairs.}
\label{fig:CACDSample}
\end{figure}

\noindent {\bf{Experiments on Cross-Age Celebrity Dataset.}} 
The CACD dataset~\cite{chen2014cross} is a face dataset for age-invariant face recognition, containing 163,446 images from 2,000 celebrities with labelled ages. It includes varying illumination, pose variation, and makeup to simulate practical scenarios. However, the CACD dataset contains some incorrectly labelled samples and some duplicate images. Following the state-of-the-art configuration~\cite{chen2015face, deng2017marginal}, we test the proposed method on a subset of CACD, called CACD-VS. The CACD-VS consists of 4,000 image pairs (2,000 positive pairs and 2,000 negative pairs) and has been carefully annotated. The 4,000 image pairs are divided into ten folds. Identities in each fold are mutually exclusive. Sample images from the CACD-VS are shown in Fig.~\ref{fig:CACDSample}. We follow the ten-folds cross-validation rule to compute the face verification rate, and compare CCL with the existing methods on this dataset. The results are listed in Table~\ref{tab:myfirsttable_cacd}. It should be noted that human performance on the CACD-VS is reported using Amazon Mechnical Turks.


\begin{table}[h]
\centering
\small
\caption{Accuracy (\%) on the CACD dataset.}
\begin{tabular}{ |@{\ }c@{\ }|@{\ }c@{\ }|@{\ }c@{\ }| } 
 \hline
 Method & Training Images & Accuracy \\
 \hline
High-Dimensional LBP~\cite{chen2013blessing}            &N/A  &81.6 \\[0.08cm]
Hidden Factor Analysis~\cite{gong2013hidden}          &N/A  &84.4 \\[0.08cm]
Cross-Age Reference Coding~\cite{chen2015face}      &N/A  &87.6 \\[0.08cm]
LF-CNNs~\cite{wen2016latent}                         &N/A  &98.5 \\[0.08cm]
Human Average~\cite{chen2015face}                   &N/A  &85.7 \\[0.08cm]
Human Voting~\cite{chen2015face}                    &N/A  &94.2 \\[0.08cm]
Centre Loss~\cite{deng2017marginal}                     &CASIA  &97.475 \\[0.08cm]
Marginal Loss~\cite{deng2017marginal}                   &4M  &98.95 \\[0.08cm]

\hline
 CCL          & CASIA (0.46M)  & \bf{99.225}\\ [0.08cm]
 CCL with AAM & CASIA (0.46M)  & \bf{99.175} \\[0.08cm]
 \hline
\end{tabular}
\label{tab:myfirsttable_cacd}
\end{table}



From Table~\ref{tab:myfirsttable_cacd}, we have the following observations:
\begin{itemize}
   \item CCL outperforms all the published results on this dataset, even surpassing the human-level performance by a clear margin. It achieves an accuracy of 99.225\%, while that of human voting is only 94.2\%. CCL with AAM has similar performance to CCL on this dataset.
   
   \item CCL shows strong robustness to age variations, though it is trained on CASIA WebFace whose data do not explicitly contain large age variation.
   
   \item The amount of our training data (0.46M) is significantly smaller than that (4M) used in Marginal Loss~\cite{deng2017marginal}, yet our method achieves better accuracy. This validates the high effectiveness of our learning model.

\end{itemize}

To some extent, the CACD-VS is a good benchmark to evaluate the robustness of an algorithm to age variation. However, we observe that the average age gap between positive pairs in CACD-VS is not large enough, and the negative pairs are randomly selected, which are not hard enough either. Therefore, we further evaluate our method on a harder dataset, the Cross-Age LFW (CALFW)~\cite{zheng2017cross}, which has larger age gap.


   
   




\begin{figure}[h]
\centering
\includegraphics[scale=0.147]{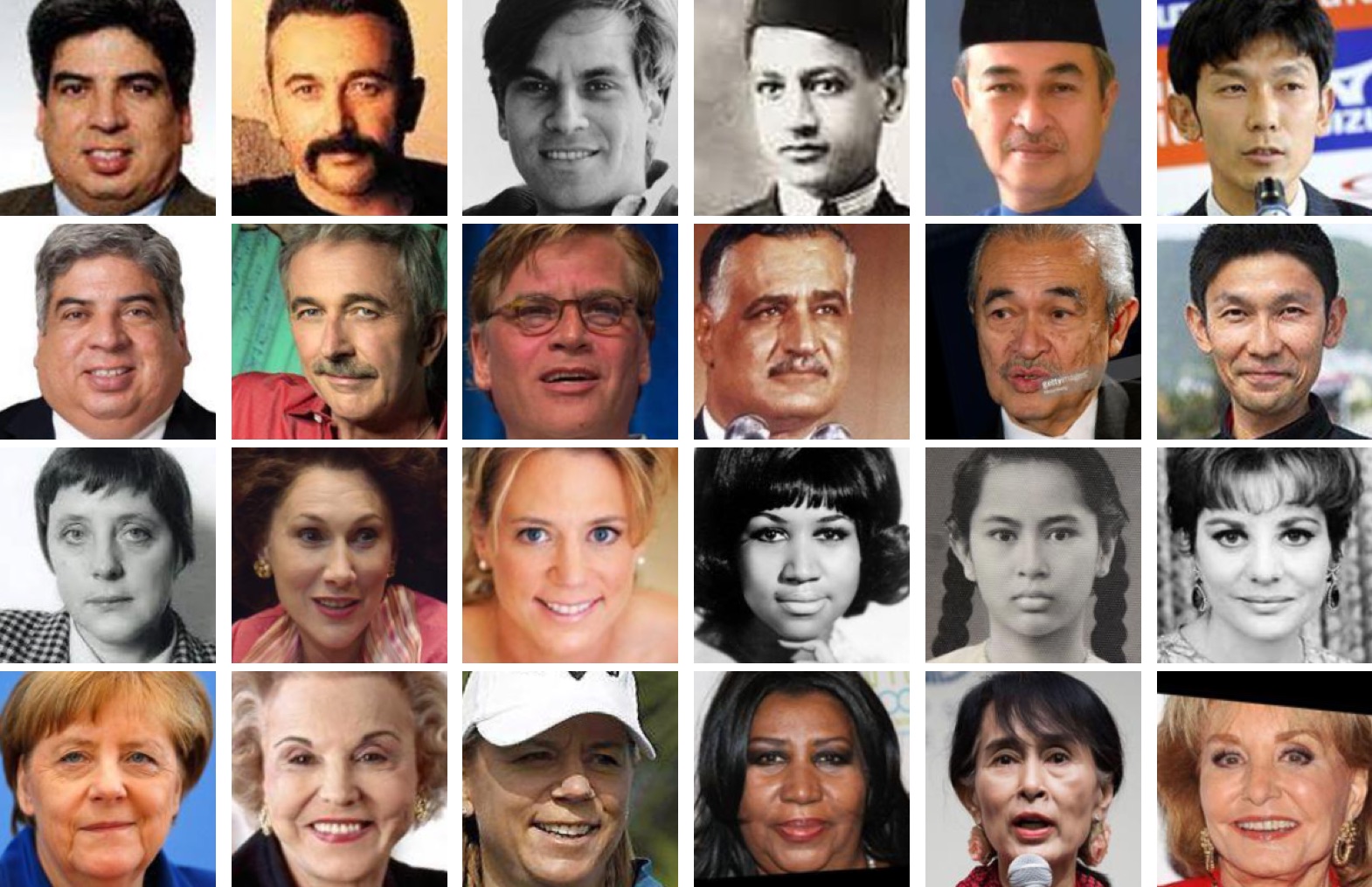}
\centering
\caption{Example positive pairs from the CALFW dataset. The first two rows have six positive male pairs and the last two rows consist of six positive female pairs.}
\label{fig:CALFWSample}
\end{figure}

\noindent {\bf{Experiments on Cross-Age LFW Dataset.}} 
The CALFW dataset~\cite{zheng2017cross} is built to evaluate face verification algorithms under large age gap. It contains 4,025 individuals with each person having 2, 3, or 4 images. Similar to the original LFW dataset, CALFW defines 10 individual subsets of image pairs. Each subset has 300 positive pairs and 300 negative pairs. These 10 subsets are constructed according to their identities to ensure that each identity only occurs in one subset. Sample image pairs are shown in Fig.~\ref{fig:CALFWSample}. One can see that there are very large age gaps between positive pairs. The age gap ranges from several years to 60 years. Large age gaps of positive pairs further increase intra-class variations.
Meanwhile, only negative pairs with the same gender and race are selected to reduce the influence of attribute difference between positive/negative pairs. Overall, this dataset is very challenging because of the large age gap and hard negative pairs.

\begin{table}[h]
\centering
\small
\caption{Accuracy (\%) on the CALFW dataset. (``+'' denotes data expansion.)}
\begin{tabular}{ |c|c|c| } 
 \hline
 Method & Training Images  &Accuracy \\
 \hline
 SVM                 &  N/A         &65.27  \\[0.08cm] 
 ITML                &  N/A         &68.82  \\[0.08cm] 
 KISSME              &  N/A         &67.87  \\[0.08cm] 
 \hline
 VGGFace~\cite{parkhi2015deep}             &  2.6M         &86.50  \\[0.08cm] 
 Noisy Softmax~\cite{chen2017noisy}        &  CASIA+      &82.52  \\ [0.08cm]
 \hline
 CCL          &  CASIA (0.46M)      &\bf{91.15} \\ [0.08cm]
 CCL with AAM &  CASIA (0.46M)      &\bf{90.83} \\[0.08cm]
 \hline
\end{tabular}
\label{tab:mycalfwtable}
\end{table}

In~\cite{zhang2017s}, Zhang et al. used Dex~\cite{rothe2015dex} to estimate the age of each image, and then calculated the age gaps in both LFW and CALFW. The average age gaps of positive pairs and negative pairs are 4.94 and 14.85, respectively, on LFW, and 16.61 and 16.14, respectively, on CALFW. We compare our method with VGGFace [38] and Noisy Softmax [59], and the results are listed in Table~\ref{tab:mycalfwtable}. One can see that our method significantly outperforms its counterparts by a large margin. It surpasses Noisy Softmax by more than 8\%. Although the verification accuracy on the original LFW is almost saturated, the performance on the CALFW is not good enough. There still has a long way to go for cross-age face recognition.

\begin{figure}[t]
\centering
\includegraphics[scale=0.14]{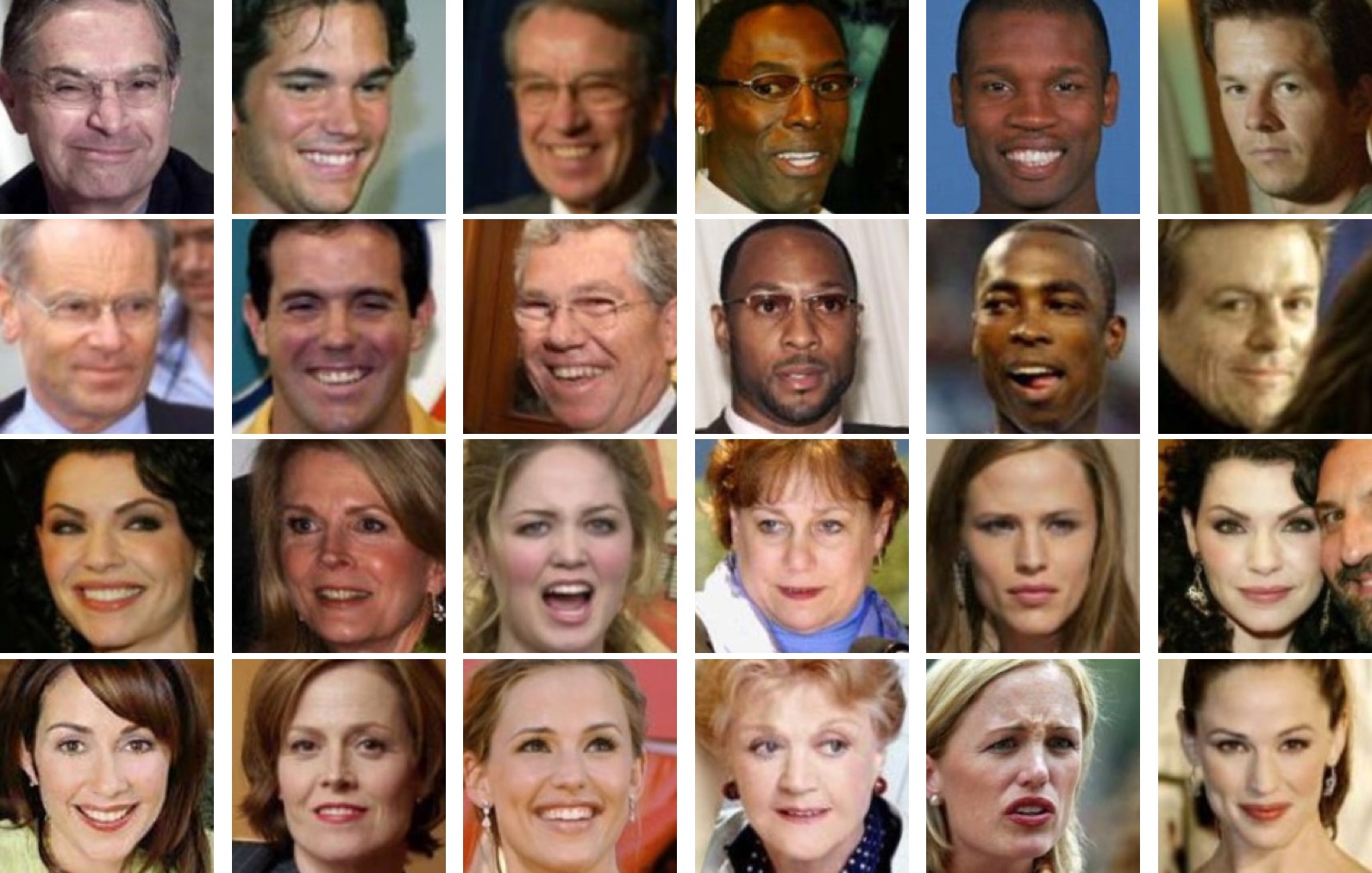}
\caption{Example negative pairs from the SLLFW dataset. The first two rows show six male negative pairs and the last two rows consist of six female negative pairs. Some pairs are even hard for human to differentiate.}
\label{fig:SLLFW}
\end{figure}

\noindent {\bf{Experiments on Similar-Looking LFW Dataset.}}
Conventional face verification addresses mainly large intra-class variations, such as pose, illumination, and expression. Zhang \etal~\cite{zheng2017cross} carefully inspected the LFW dataset. They found that the main reason for the saturated performance on LFW is that almost all negative pairs are rather easy to distinguish. They pointed out that the negative pairs were randomly selected from different individuals, and usually two randomly selected individuals will have large differences in appearance, even have different genders. In practice, however, when the gallery is large, there will be many similar-looking people as the query faces. To simulate this situation, Zhang \etal~\cite{zheng2017cross} designed the Similar-Looking LFW (SLLFW) dataset, where 3,000 similar-looking face pairs were deliberately selected from the original LFW image gallery by human crowdsourcing instead of random negative pairs selection. Fig.~\ref{fig:SLLFW} shows some negative pairs in SLLFW. One can see that the negative pairs look very similar in gender, race, age and appearance.


\begin{table}[h]
\centering
\small
\caption{Accuracy (\%) on the SLLFW dataset. (``+'' denotes data expansion.)}
\begin{tabular}{ |c|c|c| } 
 \hline
 Method & Training Images & Accuracy \\[0.08cm]
 \hline
DeepFace~\cite{taigman2014deepface}       & 0.5M  & 78.78 \\[0.08cm]
DeepID2~\cite{sun2014deep2}	       & 0.2M  & 78.25 \\[0.08cm]
VGGFace~\cite{parkhi2015deep}        & 2.6M  & 85.78 \\[0.08cm]
DCMN1~\cite{deng2017fine}          & 0.5M  & 91.00 \\[0.08cm]
Noisy Softmax~\cite{chen2017noisy}  & CASIA+  & 94.50 \\[0.08cm]
Human          & N/A   & 92.0    \\[0.08cm]
\hline
 CCL    & CASIA (0.46M)  & \bf{95.68} \\ [0.08cm]
 CCL with AAM   & CASIA (0.46M)  & \bf{96.43} \\[0.08cm]
 \hline
\end{tabular}
\label{tab:mysllfwtable}
\end{table}

We evaluate our method on SLLFW and compare it with the state-of-the-art methods. The results are listed in Table~\ref{tab:mysllfwtable}. Our CCL methods show superior performance to the other approaches. Specifically, CCL with AAM improves the Noisy Softmax by nearly 2\% using less training data. Compared with the original LFW, the performance on SLLFW drops a lot. This proves that with the increased difficulty in negative pairs, all methods will become less accurate. There is much room to improve on SLLFW.

\begin{figure}[h]
\includegraphics[scale=0.16]{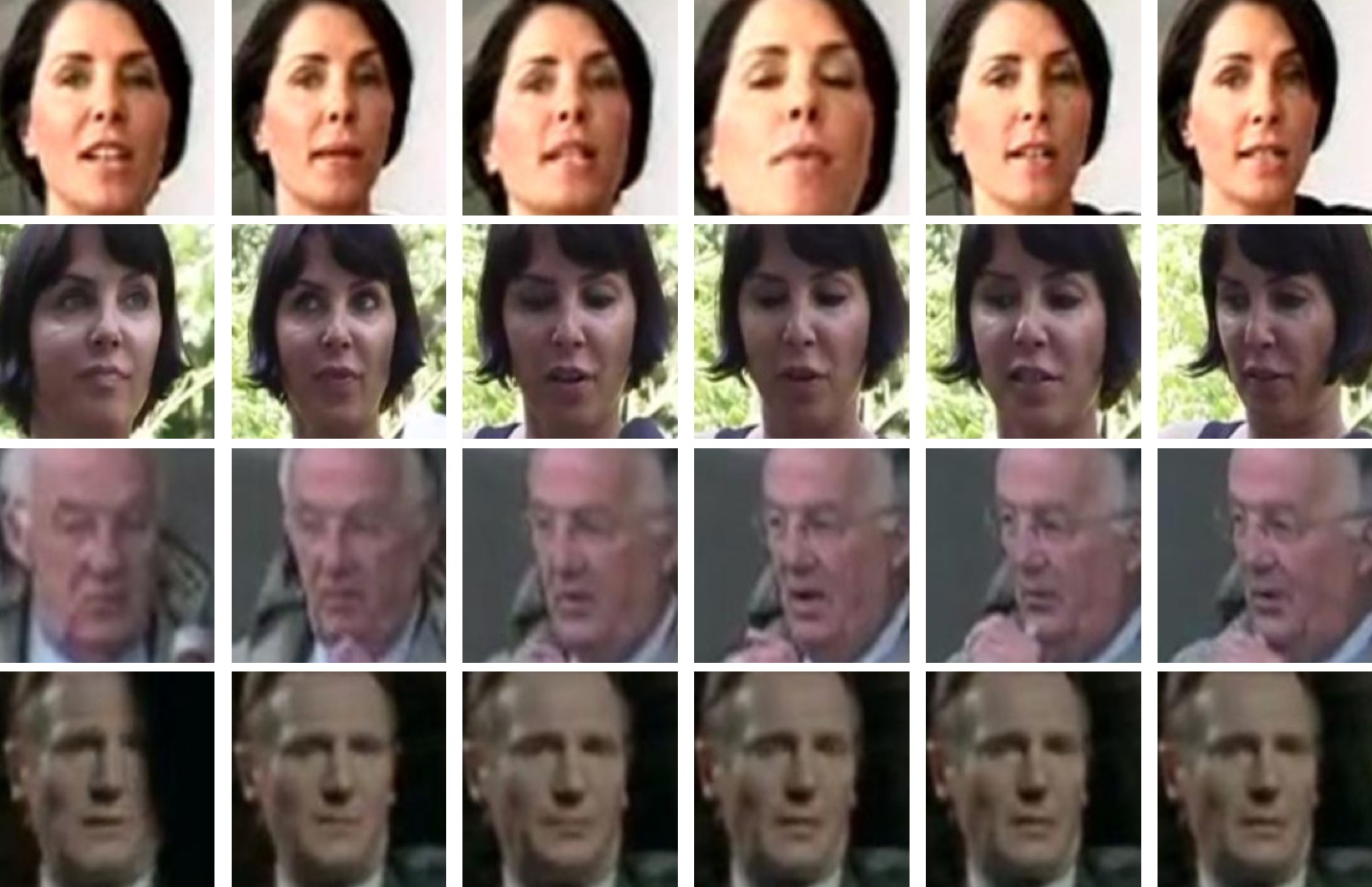}
\centering
\caption{Example pairs from the YouTube face dataset. Images in each row come from one video. The first two rows are a positive pair, and the last two rows are a negative pair.}
\label{fig:Youtube}
\end{figure}

\noindent {\bf{Experiments on YouTube Face Dataset.}}
The YTF~\cite{wolf2011face} dataset consists of 3,425 videos of 1,595 people, with an average of 2.15 videos per person. The clip durations vary from 48 frames to 6,070 frames, with an average length of 181.3 frames per clip. An essential challenge of YTF dataset is the low resolution of face images. In many videos, the size of face regions is around $40\times 40$. Some sample images are shown in Fig.~\ref{fig:Youtube}. One can see that some detected faces are very blurry. In contrast, the detected face regions in the LFW dataset are mostly larger than $120\times 120$. The YTF has 5,000 video pairs which are divided into 10 subsets. Each subset has 500 pairs: 250 positive pairs and 250 negative pairs.


\begin{table}[h]
\centering
\small
\caption{Accuracy (\%) on the YouTube Face dataset. (``*'' denotes the ensemble of 25 models.)}
\begin{tabular}{ |c|c|c| } 
 \hline
 Method & Training Images & Accuracy \\
 \hline
DeepID2+ (25)~\cite{sun2015deeply}           &0.3M*  &93.2 \\[0.08cm]
DeepFace~\cite{taigman2014deepface}          &4M  &91.4 \\[0.08cm]
FaceNet~\cite{schroff2015facenet}            &200M  &95.1 \\[0.08cm]
Center Loss~\cite{wen2016discriminative}     &0.7M  &94.9 \\[0.08cm]
Range Loss\cite{zhang2016range}              &1.5M  &93.70 \\[0.08cm]
Deep FR~\cite{parkhi2015deep}                &2.6M  &\bf{97.3} \\[0.08cm]
\hline
ReST~\cite{Wu_2017_ICCV}               &CASIA  &\bf{95.4} \\[0.08cm]
Softmax Loss                           &CASIA  &93.1 \\[0.08cm]
Softmax + Constrastive                 &CASIA  &93.5 \\[0.08cm]
$L_2$ NormFace~\cite{wang2017normface} &CASIA  &94.24 \\[0.08cm]
L-Softmax~\cite{liu2016large}          &CASIA  &94.0 \\[0.08cm]
Center Loss                            &CASIA  &94.4 \\[0.08cm]
SphereFace~\cite{liu2017sphereface}    &CASIA  &95.0 \\[0.08cm]

\hline
 CCL    &CASIA (0.46M) & 94.96 \\ [0.08cm]
 CCL with AAM   & CASIA (0.46M)  & \bf{95.28} \\[0.08cm]
 \hline
\end{tabular}
\label{tab:myytftable}
\end{table}

We follow the unrestricted with labeled outside data protocol and report the results on 5,000 video pairs in Table~\ref{tab:myytftable}. One can see that our method, trained only on the CASIA dataset, outperforms many state-of-the-art methods, including DeepFace, FaceNet and Center loss which are trained on larger dataset. Using the same CASIA training data, CCL with AAM is only slightly lower than ReST~\cite{Wu_2017_ICCV} and outperforms all the other competitors.





\noindent {\bf{Experiments on MegaFace.}}
The MegaFace dataset~\cite{kemelmacher2016megaface} is a recently released highly challenging benchmark to evaluate the performance of face recognition methods at the million scale of distractors. The MegaFace evaluation sets consist of a gallery set and a probe set. The gallery set, as a subset of Flickr photos, consists of more than one million images from more than 690K individuals. The probe set descends from two existing databases: Facescrub and FGNet. The Facescrub dataset~\cite{ng2014data}, which includes 100K photos of 530 celebrities, is available online. It has a similar number of male and female photos (55,742 photos of 265 males and 52,076 photos of 265 females) and a large variation across photos of the same individual. The probe set used in this paper consists of 3,530 images of 80 persons provided by the official MegaFace organizer.

To avoid any bias to the experimental result, we use exactly the same code to process both the probe set and the gallery set (face detection and face alignment). The experimental results are shown in Table~\ref{tab:mymegafacetable}, where ``Large'' means that the mount of used training images is larger than 500K. We can make the following observations.



\begin{table}[h]
\centering
\small
\caption{Accuracy (\%) of different methods using Facescrub as the probe set on MegaFace with 1M distractors.}
\begin{tabular}{ |c|c|c| } 
 \hline
 Method & Training Images & Accuracy \\
 \hline
 Coco Loss~\cite{liu2017rethinking}                      &Large  &\bf{76.57} \\[0.08cm]
NTechLAB - facenx large         &Large  &73.300 \\[0.05cm]
Vocord - DeepVo1                &Large  &75.127 \\[0.05cm]
Deepsense-large                 &Large  &74.049 \\[0.05cm]
Shanghai Tech                   &Large  &74.799 \\[0.05cm]
Google - FaceNet v8             &Large  &70.496 \\[0.05cm]
Beijing FaceAll\_Norm\_1600     &Large  &64.804 \\[0.05cm]
Beijing FaceAll\_1600           &Large  &63.977 \\[0.05cm]
\hline
Deepsense-small                 &Small  &70.983 \\[0.05cm]
SIAT\_MMLAB                     &Small  &65.233 \\[0.05cm]
barebones FR                    &Small  &59.036 \\[0.05cm]
NTechLAB-facenx\_small          &Small  &66.366 \\[0.05cm]
3DiVi Company-tdvm6             &Small  &36.927 \\[0.05cm]
\hline
Softmax Loss~\cite{liu2017sphereface}                    &Small  &54.855 \\[0.05cm]
Softmax+Contrastive~\cite{liu2017sphereface}             &Small  &65.219 \\[0.05cm]
Triplet Loss~\cite{liu2017sphereface}                    &Small  &64.797 \\[0.05cm]
L-Softmax Loss~\cite{liu2017sphereface}                  &Small  &67.128 \\[0.05cm]
Softmax+Center Loss~\cite{liu2017sphereface}             &Small  &65.494 \\[0.05cm]
SphereFace~\cite{liu2017sphereface}                   &Small  &\bf{72.729} \\[0.05cm]

\hline
 CCL    & Small  & 72.572 \\ [0.05cm]
 CCL with AAM   & Small  & \bf{73.743} \\[0.08cm]
 \hline
\end{tabular}
\label{tab:mymegafacetable}
\end{table}

\begin{figure*}[t]
\centering
\includegraphics[scale=0.50]{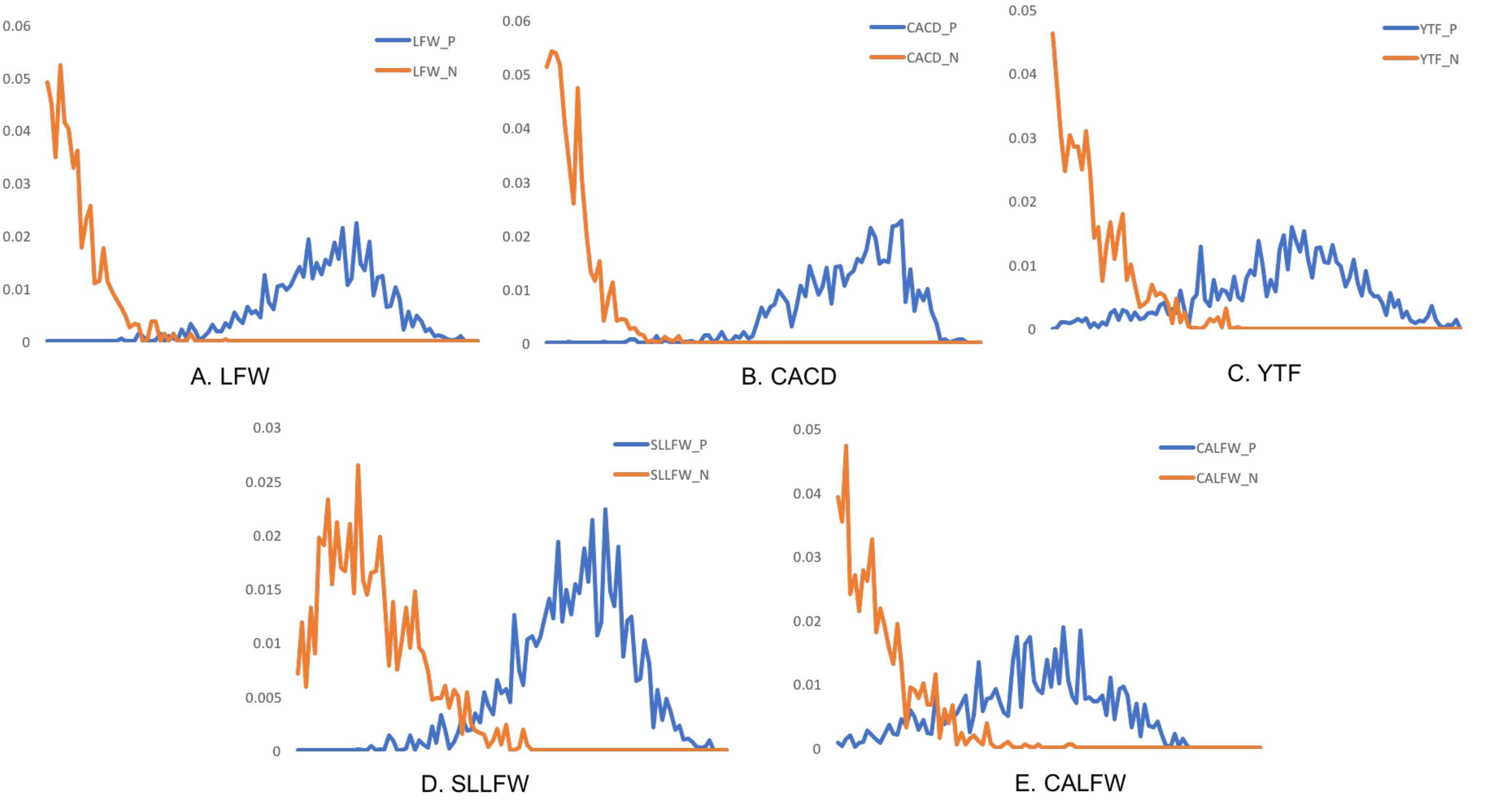}
\centering
\caption{Histogram distributions of similarity scores on LFW, CACD, YTF, SLLFW, and CALFW data sets. Similarities of positive pairs are marked blue color and similarities of positive pairs are marked orange color.}
\label{fig:5LFW}
\end{figure*}

\begin{figure*}[t]
\centering
\includegraphics[scale=0.24]{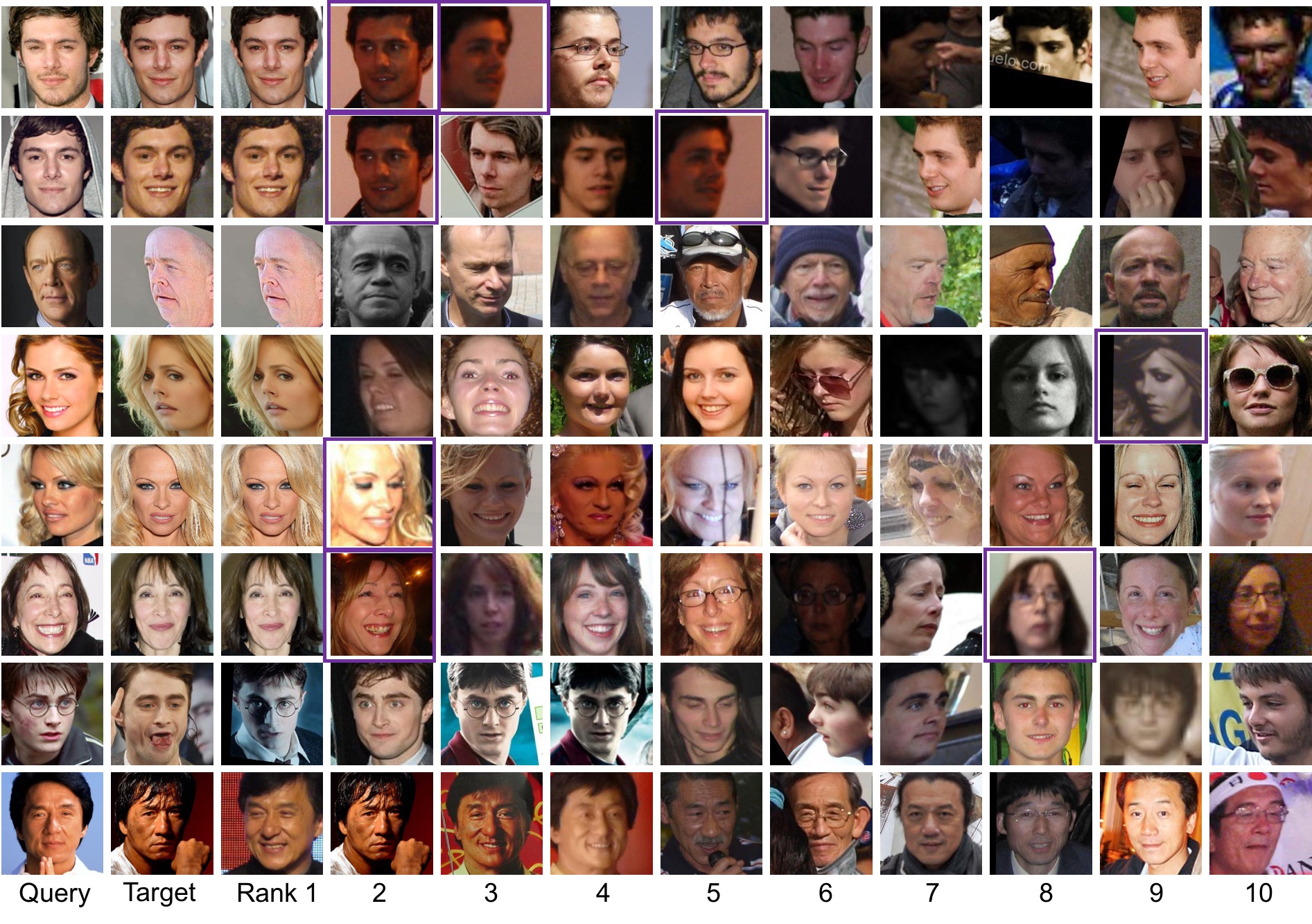}
\caption{Retrieval results on MegaFace. The first column is the query image, and the second column is the target image. The last ten columns are the top ten retrieval results.}
\label{fig:megaface222}
\end{figure*}

\begin{itemize}
    \item Among the methods which are trained on small scale data, the proposed CCL with AAM achieves the best accuracy. It largely outperforms Softmax loss, L-Softmax loss, Triplet loss, and Center Loss + Softmax loss. Without using AAM, CCL achieves similar performance (72.572\%) to SphereFace (72.729\%) with large angular margin $(m = 4)$. By using AAM, CCL outperforms SphereFace by about 1\%.
    \item The proposed CCL with AAM also outperforms many methods trained on large scale data. The Coco Loss [41] method achieves the accuracy of 76.57\%, about 2.8\% higher than CCL with AAM, but it uses  half MS-Celebrity (about 2M images) in training, while our methods uses only 0.46M images from 10,575 classes in training. 
\end{itemize}




\subsection{Discussions}
\label{sec:4.4}
\noindent {\bf{Discussions about LFW, CACD, YTF, SLLFW and CALFW data sets.}}
From our experiments in Section~\ref{sec:4.3}, one can see that the face recognition performance on different datasets varies much. We first compare the LFW, CACD, CALFW, SLLFW and YTF datasets by computing the statistics of similarity scores of positive pairs and negative pairs, respectively. The similarity distributions on the five datasets are drawn in Fig.~\ref{fig:5LFW}. We can see that there are clear boundaries between the similarity distribution of positive pairs and the similarity distribution of negative pairs on the LFW and CACD datasets, which explains why the performance on these two datasets is already saturated. The two distribution curves have some overlaps on the YTF and SLLFW datasets, which implies that more efforts should be made to further improve the performance on these two datasets. For the CALFW dataset, the two distribution curves overlap much. This is mainly caused by the large age gap of face images from the same subject. Some positive pairs have lower similarity scores, and few positive pairs have a similarity score close to 1.0. CALFW is a challenging dataset to evaluate the cross-age face recognition algorithms.



\noindent {\bf{Discussions about MegaFace dataset.}} 
Compare with the above 5 datasets, MegaFace is much bigger in scale and it also has a different yet more challenging test protocol. To visualize the results on MegaFace, we conduct a retrieval experiment. Each time, we select two images of one subject from the Facascrub probe set (3,540 images), one used as ``Query'' image and the other one used as ``Target'' image. The target image is mixed with the gallery set which has 1,027,060 images. We use the query image to retrieval the ten most similar faces from the 1,027,061 images. Some retrieval results are shown in Fig.~\ref{fig:megaface222}.

Some observations can be made from the retrieval results in Fig.~\ref{fig:megaface222}. In particular, it is very hard to decide whether some retrieval results, marked with purple rectangles, are truly from the same people as the query images. Since the gallery set is very large and the probe set is collected all from celebrities, the gallery set actually contain many images from the identities appearing in the probe set. Take the last two rows for example, the query images are from ``Daniel Jacob Radcliffe'' and ``Jackie Chan''; however, we can find that the gallery set contain multiple images coming from the same identities. According to the ``Rank 1'' criterion, the retrieval results for ``Daniel Jacob Radcliffe'' and ``Jackie Chan'' are wrong, but we can observe that the top 4 results for ``Daniel Jacob Radcliffe'' and ``Jackie Chan'' are actually correct.
The distractors in the gallery set do not exclude the images with the same subject as the query subject. Such a fact will affect the veracity of the accuracy criterion. 




\section{Conclusion}
This paper presented a simple yet effective face feature learning method. With the help of deep convolutional neural networks (CNNs), we argued that a good face feature learner should push the face samples dispersedly distributed across the coordinate space centering on the origin so that the angles between different classes can be enlarged. Meanwhile, the classification vectors should lie on a hypersphere space to remove the influence of their varying magnitudes. To achieve this goal, we normalized the classification vector by its $L_2$ norm, and centralized each dimension of the face feature to zero mean with unit variance. An adaptive angular margin was also defined to further enhance the separability of neighboring classes. The proposed method, namely centralized coordinate learning (CCL), was trained on the CASIA Webface dataset, which has only 0.46M face images from about 10K persons. Extensive experiments on six benchmarks, including LFW, CACD, SLLFW, CALFW, YouTube Face and MegaFace, were conducted. CCL consistently exhibits competitive performance on all the six databases. It also outperforms many state-of-the-art models which are trained on much larger datasets.

{
\Large
\bibliographystyle{IEEEtran}
\bibliography{egbib.bib}
}

\ifCLASSOPTIONcaptionsoff
  \newpage
\fi

\end{document}